\documentclass[lettersize,journal]{IEEEtran}
\usepackage{amsmath,amsfonts}
\usepackage{array}
\usepackage[caption=false,font=normalsize,labelfont=sf,textfont=sf]{subfig}
\usepackage{textcomp}
\usepackage{stfloats}
\usepackage{url}
\usepackage{verbatim}
\usepackage{graphicx}
\usepackage{booktabs}
\usepackage{adjustbox}
\usepackage{tabularx}
\usepackage{cite}
\begin{document}

\title{OV-PEAR: Observation-Verified Pose-Scale Estimation and Reconstruction for Unknown Industrial Manipulation Objects}

\author{Guoqin Tang, Qingxuan Jia, Yichen Tan, Zihang Lv, Ruiquan Wang, Zeyuan Huang, and Gang Chen\textsuperscript{*}
	\thanks{*Corresponding author (e-mail: tgq@bupt.edu.cn).}
	\thanks{All authors are with the School of Intelligent Engineering and Automation, Beijing University of Posts and Telecommunications, Beijing 100876, China.}
}

\maketitle

\begin{abstract}
Fine industrial robotic manipulation requires 6D pose, scale, and an operation-ready 3D model. Existing industrial pose estimation relies on CAD models, object-specific training data, or stable workcells, limiting use with unseen objects; zero-shot methods suffer from single-view occlusion, generative bias, and scale uncertainty. We present OV-PEAR, an observation-verified framework for 3D reconstruction and pose-scale estimation of unknown industrial objects. Its three-stage object-centric pipeline: (1) segments RGB-D instances and generates object-level 3D hypotheses; (2) anchors them to the robot frame via appearance-aware view retrieval, explicit scale search, and visibility-sensitive registration; (3) back-projects multi-view RGB-D evidence into the object-canonical frame to verify voxel occupancy, suppress hallucinations, and produce reusable trusted object models. On BOP-Industrial, OV-PEAR achieves 0.657 average AR and improves pose and voxel-level geometric consistency in real scenes; real-robot connector assembly reaches 99.67\% grasp and 84.00\% assembly success. The model library and pose-scale log enable short-term object-level state maintenance at 0.02 ms/frame.

\end{abstract}


\section{Introduction}
\IEEEPARstart{I}{ndustrial} robotic applications typically include both mobile robotic tasks, such as inspection and environment exploration, and manipulation tasks, such as assembly, sorting, and palletizing. Compared with navigation tasks, which mainly rely on scene-level understanding and path planning, manipulation tasks require more precise object-centric perception. To generate executable grasping, insertion, or assembly motions, a robot must not only localize the target object, but also estimate its accurate 6D pose, physical scale, and contact-relevant local geometry, so that the end-effector pose and manipulator configuration can be planned accordingly~\cite{zhuang2023instance,mahler2019ambidextrous,chen2025embodiedai}. This requirement is particularly critical in industrial workcells, where manipulated objects may vary across product models, be updated frequently, or lack readily available CAD models and training samples. In addition, inter-object occlusion, placement perturbations, and workcell reconfiguration are common in industrial environments, forcing robots to operate under continuously changing object categories, poses, and visibility conditions. Therefore, achieving operation-ready high-precision perception for unseen manipulation objects in variable industrial scenes is a key prerequisite for autonomous industrial robotic manipulation.

Most existing pose-perception methods for industrial robots are built on the assumption of known objects and relatively stable workcell configurations, making it difficult to simultaneously satisfy the requirements of unseen objects, scene variability, and fine manipulation accuracy. Traditional pose-estimation methods typically rely on pre-built target models, controlled imaging conditions, and stable scene layouts, and localize objects through 2D feature extraction, matching, and geometric solving~\cite{marullo2022sixd,yuanwei2024review}. Deep learning-based methods have substantially improved feature representation under complex conditions, but their performance often depends on large-scale annotated data collected for specific objects, workcells, and operating conditions~\cite{apostolopoulos2022industrial,akar2022synthetic}. These methods can be effective in standardized production lines; however, their generalization and estimation accuracy often degrade when new objects are introduced, occlusion becomes stronger, backgrounds change, or workstations are reconfigured. For unseen manipulation objects without available templates, CAD models, or training samples, existing methods still struggle to provide high-accuracy pose estimates that can directly support fine manipulation without retraining. This motivates the need for a high-precision pose-perception method that can handle unseen objects and scene variations without scene-specific training.

Accurate pose estimation alone, however, is still insufficient for fine manipulation. For grasping, insertion, and assembly, robots also require geometric information about the target object in order to reason about contact regions, kinematic constraints, and feasible manipulation poses. This requirement is particularly important for unseen manipulation objects. Recently, zero-shot reconstruction methods that exploit priors from visual foundation models have provided a promising direction for estimating the structure of unknown objects. Representative methods such as SAM3D can leverage semantic and shape priors from pretrained models to rapidly generate complete object-level geometry from a single image or a small number of observations, showing strong generality in open environments~\cite{zhou2024comprehensive,sam3d2025,fan2025vlm3r}. Nevertheless, their structural predictions are largely driven by learned priors rather than continuously verified observations. Under occlusion, incomplete sensing, and cluttered industrial backgrounds, such predictions may contain geometric hallucinations or local structural deviations. Moreover, most existing methods rely on single-shot prediction or one-time fusion, and lack a mechanism for continuously incorporating multi-view evidence to revise the target model. Therefore, for unseen manipulation objects, a key challenge is how to exploit model priors for generalization while using incremental observations to suppress prior-induced errors and obtain a trusted 3D object model for industrial robotic manipulation.

The above challenges can be illustrated by a typical industrial assembly scenario. On a cluttered assembly workbench, multiple connector types, such as USB, RJ45, and HAN10 connectors, may appear simultaneously. These objects can have similar external shapes but different assembly constraints. Existing industrial model libraries or historical templates may cover only a subset of standard objects, while newly introduced parts remain unseen to the robot. At the same time, large occluders may prevent a single camera view from capturing complete pose and structural information. Under such conditions, a robot that relies only on pose estimators trained for known objects may fail to identify and localize unseen connectors accurately. Conversely, directly using a one-shot generated 3D model may introduce structural distortions due to partial occlusion and prior-induced bias, thereby reducing the reliability of downstream insertion. To execute such manipulation tasks, the robot must actively acquire multi-view RGB-D observations with an eye-in-hand camera and update both the pose estimate and the geometric model of the target object as new observations become available.

To address these issues, we propose OV-PEAR, an observation-verified zero-shot framework for pose-scale perception and 3D reconstruction of unseen industrial manipulation objects. The framework is designed to estimate both an object model and an accurate pose-scale state without requiring a target CAD model, object-specific training data, or scene-specific retraining, while also providing an interface for subsequent object-model reuse. Unlike methods that directly treat a single-view generated result as the final operation model, OV-PEAR uses the generative 3D object prior as an initial object hypothesis. It then anchors this hypothesis to the robot frame and verifies its geometry using robot-acquired RGB-D observations, producing a trusted object-level perception representation for robotic manipulation.

Specifically, OV-PEAR first constructs an initial geometric hypothesis for an unseen object using a single-view generative 3D model. It then instantiates the object model in the robot coordinate frame through appearance-aware view retrieval, explicit scale search, and visibility-sensitive geometric registration. Furthermore, multi-view RGB-D evidence is used to verify and update the generated voxel occupancy, suppress hallucinated structures, and improve consistency between the reconstructed object model and real observations. Based on the verified models, OV-PEAR organizes object geometry into a reusable object model library. During subsequent continuous observations, the system updates only the pose, scale, and visibility state of each object instance, enabling compact object-level state records for short-term scene replay. We evaluate the proposed framework on public industrial benchmarks, real workbench scenes with unseen objects, fine insertion initialization tasks, and short-term continuous-frame scene replay experiments, demonstrating its effectiveness in zero-shot pose estimation, object-model refinement, robotic manipulation initialization, and reusable object-level scene representation.

The main contributions of this paper are summarized as follows:
\begin{enumerate}
	\item We propose OV-PEAR, an observation-verified framework that converts generative 3D priors into robot-usable object models and pose-scale states for unseen industrial manipulation objects without target CAD models or object-specific training.
	\item We develop a pose-scale anchoring method that integrates appearance-aware view retrieval, explicit scale search, and visibility-sensitive registration to instantiate generated object hypotheses in the robot frame.
	\item We introduce an observation-verified reconstruction scheme that uses multi-view RGB-D evidence to update voxel occupancy, suppress hallucinated geometry, and improve the trustworthiness of reconstructed object models.
	\item We validate OV-PEAR on public industrial benchmarks, real workbench scenes, connector insertion tasks, and continuous-frame dynamic scenes, demonstrating its effectiveness for unseen-object perception, manipulation initialization, and short-term object-level scene maintenance.
\end{enumerate}

\section{Related Work}
\subsection{6D Pose Estimation}

6D pose estimation is a fundamental problem for fine industrial manipulation, aiming to recover the position and orientation of a target object under clutter, occlusion, and sensor noise. Existing methods can be broadly grouped into model-based, data-driven, and zero-shot paradigms, each of which faces different limitations in flexible industrial deployment.

Classical model-based methods estimate object pose by matching pre-built templates or local geometric descriptors to online observations. Template-based approaches, such as LineMod~\cite{hinterstoisser2011multimodal}, match the scene image against rendered views of a known 3D model, while feature-based methods, such as PFH~\cite{rusu2008aligning} and FPFH~\cite{rusu2009fast}, establish point-cloud correspondences and often refine the result using ICP. These methods can be accurate and efficient when the target model is available and the observation is sufficiently complete. However, their dependence on CAD models, reliable features, and limited occlusion makes them less suitable for cluttered industrial workcells containing unseen objects.

Deep learning-based methods have improved pose estimation by learning stronger visual and geometric representations. RGB methods such as PoseCNN~\cite{xiang2018posecnn}, SSD-6D~\cite{kehl2017ssd6d}, and YOLO-6D~\cite{tekin2018realtime} regress poses or keypoints from images, while RGB-D methods such as PVN3D~\cite{he2020pvn3d} and G2L-Net~\cite{chen2020g2lnet} incorporate depth information to strengthen geometric constraints. Despite their strong performance on standard benchmarks, these methods typically require object-specific or scene-specific annotated data. Their accuracy can degrade when new parts are introduced, workstations are reconfigured, or background and occlusion patterns change, increasing the cost of deployment in flexible manufacturing.

Recent zero-shot pose-estimation methods, represented by SAM-6D~\cite{lin2024sam6d}, exploit priors from visual foundation models to infer poses for unseen objects without object-specific training. This direction is promising for rapid industrial changeover, but existing methods are still largely constrained by single-view observations and are often more reliable as coarse pose initializers than as high-precision estimators for fine manipulation. Under severe occlusion, scale uncertainty, and strict spatial-alignment requirements, it remains challenging to achieve both zero-shot generalization and operation-level pose accuracy. This motivates a pose-scale anchoring mechanism that can instantiate generated object hypotheses in the robot frame under partial observations.

\subsection{3D Reconstruction}

Fine manipulation also requires operation-ready object geometry, since grasping, insertion, and assembly depend on contact regions, local constraints, and collision geometry. Existing 3D reconstruction methods include classical multi-view reconstruction, learning-based reconstruction, and zero-shot generative reconstruction, but none of these directly provides a reliable and continuously verified object model for unseen industrial manipulation objects.

Classical SfM and MVS methods, such as global or incremental SfM~\cite{moulon2013global,yin2020incremental}, COLMAP~\cite{schonberger2016structure}, and PMVS~\cite{furukawa2010accurate}, recover 3D structure by enforcing multi-view geometric consistency. These methods can produce high-quality reconstructions when sufficient viewpoints, texture, and camera motion are available. However, their computational cost and dependence on favorable viewing conditions limit their applicability to rapid object modeling in cluttered industrial workcells, where the target may be partially occluded or observable only from constrained viewpoints.

Learning-based reconstruction methods, including MVSNet~\cite{yao2018mvsnet}, Pixel2Mesh~\cite{wang2018pixel2mesh}, and Neuralangelo~\cite{li2023neuralangelo}, improve reconstruction efficiency and surface quality by learning priors from data. Nevertheless, their performance is still tied to training distributions, and additional data collection or retraining may be required when object categories, workcell layouts, or sensing conditions change. More recently, zero-shot reconstruction methods such as VGGT~\cite{wang2025vggt} and SAM3D~\cite{sam3d2025} have used foundation-model priors to generate 3D object geometry from limited observations, while inverse-neural-rendering work has explored generalizable and interpretable 3D tracking under distribution shifts~\cite{ost2025generalizable}. These methods are attractive for unseen-object modeling, but their outputs are still driven by learned shape priors and may contain hallucinated structures, scale deviations, or local geometric errors under occlusion and incomplete sensing.

Overall, existing pose-estimation and reconstruction methods address complementary parts of the manipulation perception problem, but they do not fully solve the joint requirement of accurate pose-scale anchoring and trustworthy geometry construction for unseen industrial objects. What is still missing is a mechanism that uses generative 3D priors to bootstrap object modeling, anchors the generated hypothesis to the robot frame, and then verifies and corrects the geometry with multi-view RGB-D evidence. OV-PEAR is designed to fill this gap through pose-scale anchoring and observation-verified voxel updating.

\section{Methodology}

OV-PEAR is designed for perception of unknown industrial manipulation objects without a target CAD model, object-specific training data, or scene-specific retraining. Given RGB-D observations, the goal is to obtain an observation-verified 3D object model, a pose-scale state in the robot coordinate frame, and an object-level representation that can be reused in subsequent manipulation and short-term state maintenance. The method consists of three stages. First, a pretrained single-view voxel generator produces initial geometric hypotheses for unknown object instances. Second, pose-scale anchoring instantiates each generated object model in the robot frame. Third, multi-view RGB-D evidence verifies and corrects voxel occupancy, yielding a trusted object model that is inserted into an object-model library. This section describes the object-centric data flow, the pose-scale anchoring module, and the observation-verified multi-view reconstruction module.

\subsection{OV-PEAR Architecture and Object-Centric Data Flow}

\begin{figure*}[!t]
	\centering
	\includegraphics[width=\textwidth]{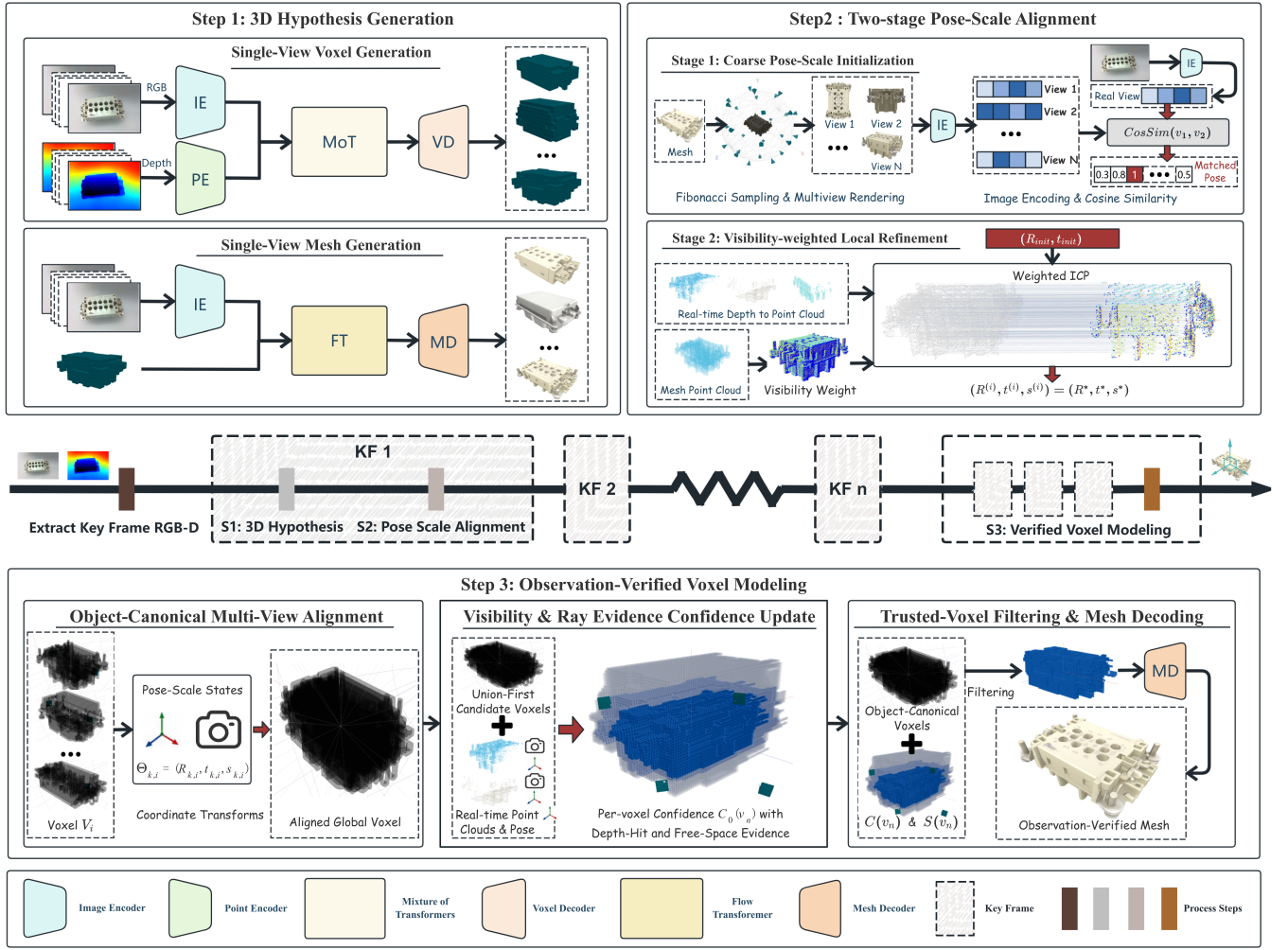}
	\caption{Timeline-centered overview of the proposed OV-PEAR framework. After key-frame RGB-D observations are extracted, each key frame first supports 3D hypothesis generation and two-stage pose-scale alignment. Evidence accumulated across multiple key frames is then integrated for observation-verified voxel modeling, trusted-voxel filtering, and mesh decoding.}
	\label{fig:architecture}
\end{figure*}

As shown in Fig.~\ref{fig:architecture}, OV-PEAR follows an object-centric perception pipeline. 

OV-PEAR starts from scene-level RGB-D observations rather than from pre-cropped object images. Let the robot acquire RGB-D observations from $N$ camera poses,

\[
X=\left\{(I_i,D_i,T_i^c)\right\}_{i=1}^{N},
\]

where $I_i$ and $D_i$ denote the RGB image and depth map of the $i$th view, respectively, and $T_i^c$ denotes the extrinsic transformation from the camera coordinate frame of that view to the robot base frame or the world frame. These observations can be collected by a wrist-mounted RGB-D camera moving along a planned trajectory, or by a calibrated multi-camera system.

For each view, an instance segmentation front end first detects candidate object masks in the scene:

\[
B_i=\operatorname{Seg}(I_i)=\left\{B_{i,j}\right\}_{j=1}^{J_i},
\]

where $B_{i,j}$ is the pixel-level mask of the $j$th candidate instance in view $i$, and $J_i$ is the number of candidate instances in that view. In our implementation, $\operatorname{Seg}(\cdot)$ can be instantiated by SAM3 or another open-vocabulary instance segmentation model derived from the Segment Anything family~\cite{kirillov2023segment}. Its role is to convert the full scene image into object-level candidate observations, rather than to serve as the core reconstruction module.

The system then associates candidate masks across views or consecutive frames using 2D appearance features, depth-projected positions, 3D spatial consistency, and task-level constraints. This gives the multi-view observation set of the $k$th target object:

\[
O_k=\operatorname{Assoc}\!\left(\left\{(B_i,D_i,T_i^c,I_i)\right\}_{i=1}^{N}\right).
\]

For a valid observation of object $k$ in view $i$, let its mask be $B_{k,i}$. The local RGB observation is

\[
I_{k,i}=I_i\odot B_{k,i},
\]

and the corresponding local 3D point cloud is

\[
P_{k,i}=T_i^c\,\Pi^{-1}(D_i,B_{k,i}),
\]

where $\Pi^{-1}(\cdot)$ denotes back-projection from the depth map and instance mask using the camera intrinsics. The original scene-level input is thereby converted into the object-centric observation sequence

\[
X_k=\left\{(I_{k,i},B_{k,i},P_{k,i},T_i^c)\right\}_{i\in\mathcal{I}_k},
\]

where $\mathcal{I}_k$ is the set of view indices in which object $k$ is validly observed.

OV-PEAR does not require a target object to remain static over the entire task sequence. It only assumes rigid shape consistency during one modeling or update window, and it requires object correspondences to be established across views or frames through instance segmentation, appearance matching, or state association. If an object translates or rotates, the estimated pose-scale state of each frame is used to compensate local RGB-D evidence into a common object-canonical coordinate frame. Multi-view geometric fusion therefore relies on object-level pose compensation, not on a globally static scene assumption. Non-rigid deformation and topological changes are outside the scope of this work.

For each valid view of object $k$, the single-view voxel generation module $G(\cdot)$ produces a voxel hypothesis from the local RGB/mask observation:

\[
V_{k,i}=G(I_{k,i},B_{k,i}),\qquad i\in\mathcal{I}_k.
\]

This voxel hypothesis is treated only as an initial geometric candidate for an unknown object. It is not used directly as the final operation model, since single-view generation can be affected by occlusion, segmentation errors, degenerate viewpoints, generative-prior bias, and scale ambiguity. OV-PEAR therefore instantiates and corrects the hypothesis through pose-scale anchoring and multi-view observation verification.

The system selects a reference view $r\in\mathcal{I}_k$ from the valid observation set and decodes its voxel representation $V_{k,r}$ into a reference object model:

\[
M_{k,r}=D(V_{k,r}).
\]

The reference model $M_{k,r}$ is not the final optimized model. It is used to build candidate view templates, extract appearance and geometric features, and provide initial object geometry for pose-scale anchoring. Based on this reference model, the system constructs a candidate template or feature library:

\[
\mathcal{T}_k=\operatorname{Temp}(M_{k,r}),
\]

where $\mathcal{T}_k$ denotes the set of templates rendered or extracted from the reference model. For each observed view, template matching or appearance-feature retrieval gives an initial view hypothesis:

\[
q_{k,i}^{0}=\operatorname{Match}(I_{k,i},B_{k,i},\mathcal{T}_k).
\]

This initial view hypothesis reduces the canonical-view ambiguity of the generated object model and initializes the subsequent pose-scale anchoring step.

Given the local RGB-D observation, the reference object model, the initial view hypothesis, and the camera extrinsic parameters, the pose-scale anchoring module estimates a similarity transformation for object $k$ in view $i$:

\[
\Theta_{k,i}=(R_{k,i},t_{k,i},s_{k,i})
=A(I_{k,i},B_{k,i},P_{k,i},M_{k,r},q_{k,i}^{0},T_i^c),
\]

where $R_{k,i}\in SO(3)$ is the rotation matrix, $t_{k,i}\in\mathbb{R}^{3}$ is the translation vector, and $s_{k,i}\in\mathbb{R}_{+}$ is the scale factor. To avoid coordinate ambiguity, $\Theta_{k,i}$ is defined as the pose-scale transformation from the object-canonical coordinate frame of object $k$ to the robot base frame or the world frame. The internal mechanism of $A(\cdot)$ is described in the next subsection.

After obtaining $\Theta_{k,i}$, OV-PEAR maps generated voxel hypotheses and real depth evidence into the same object-canonical coordinate frame. For the generated voxels, the aligned voxel hypothesis is

\[
\bar V_{k,i}=W_{\mathrm{vox}}(V_{k,i};\Theta_{k,i}),
\]

where $W_{\mathrm{vox}}(\cdot)$ denotes the voxel-space transformation that aligns the generated voxel hypothesis of view $i$ to the object-canonical frame. For the real depth point cloud and camera center, the inverse transformation $\Theta_{k,i}^{-1}$ compensates the evidence from the robot base frame or world frame to the object-canonical frame:

\[
\bar P_{k,i}=W_{\mathrm{dep}}(P_{k,i};\Theta_{k,i}^{-1}),\qquad
\bar o_{k,i}=W_{\mathrm{dep}}(o_i;\Theta_{k,i}^{-1}),
\]

where $o_i$ is the camera center of view $i$. After this step, local geometric evidence collected from different camera poses, different observation times, and even rigid object motion is expressed in the object coordinate frame.

In the object-canonical coordinate frame, OV-PEAR first aggregates multi-view voxel hypotheses into a candidate set:

\[
V_{k,\mathrm{cand}}=\bigcup_{i\in\mathcal{I}_k}\bar V_{k,i}.
\]

This union-first candidate set preserves structural recall from multi-view generated hypotheses and prevents an early, irreversible deletion caused by a locally incomplete low-quality view. The system then verifies and carves the candidate voxels using real RGB-D depth evidence, camera centers, visibility relations, and multi-view support:

\[
V_{k,\mathrm{trust}}
=U_{\mathrm{carve}}\!\left(
V_{k,\mathrm{cand}},
\left\{(\bar P_{k,i},\bar o_{k,i},B_{k,i},\Theta_{k,i})\right\}_{i\in\mathcal{I}_k}
\right).
\]

Here, $U_{\mathrm{carve}}(\cdot)$ denotes the trusted-voxel selection module that combines depth-surface hits, free-space ray carving, and multi-view support. The output $V_{k,\mathrm{trust}}$ is the set of voxels verified by real geometric evidence. The detailed design of this module is given in the observation-verification subsection.

Finally, the system generates an optimized object mesh by combining the trusted voxels, the reference RGB image, and the reference mask:

\[
M_k^\ast=D_{\mathrm{mesh}}(V_{k,\mathrm{trust}},I_{k,r},B_{k,r}).
\]

Here, $V_{k,\mathrm{trust}}$ provides trusted geometric support in the object-canonical coordinate frame, while $I_{k,r}$ and $B_{k,r}$ provide appearance, boundary, and color information. The optimized object model $M_k^\ast$ is inserted into the object-model library:

\[
L\leftarrow L\cup\{M_k^\ast\}.
\]

The final object model in OV-PEAR is therefore neither a direct reuse of a single-view generated result nor a simple extraction of a coarse voxel surface. It is constrained jointly by generative candidate geometry, pose-scale anchoring, real depth carving, and RGB/mask appearance information.

Once the model library has been established, subsequent frames do not require repeated full object-model construction for previously modeled objects. For the $t$th input frame, the system obtains current instance masks and uses library-template matching and state association to determine the correspondence between each object instance and the model library:

\[
(\ell_{k,t},\Theta_{k,t})
=\operatorname{LibMatch}(I_t,D_t,B_{k,t}^{\mathrm{obs}},L),
\]

where $\ell_{k,t}$ is the model-library index of object instance $k$, $\Theta_{k,t}$ is the pose-scale state estimated for the current frame, and $B_{k,t}^{\mathrm{obs}}$ is the observed instance mask. In this paper, $\operatorname{LibMatch}(\cdot)$ is used as the system interface for object-model reuse. Its effectiveness is evaluated through the short-term continuous-frame state maintenance experiments, while the core modeling and anchoring mechanisms are detailed below.

The system further maintains an object-level state record:

\[
S_t=\left\{
(\mathrm{id}_{k,t},\ell_{k,t},T_{k,t},s_{k,t},c_{k,t},v_{k,t})
\right\}_{k=1}^{N_t},
\]

where $\mathrm{id}_{k,t}$ is the object-instance identifier, $\ell_{k,t}$ is the model-library index, $T_{k,t}$ and $s_{k,t}$ are the pose and scale of the object in frame $t$, $c_{k,t}$ is the state confidence, and $v_{k,t}$ is the visibility state. This record decouples object geometry from temporal state. Short-term state maintenance can therefore reuse the shared object model in the library and update only the pose, scale, identity, and visibility of each instance, without regenerating a full 3D model at every frame.

Overall, the object-centric data flow of OV-PEAR converts scene-level RGB-D observations into an object-level model library and state log. Instance segmentation and object association create object-level observation units. The single-view voxel generator provides initial geometric candidates for unknown objects. Template matching and pose-scale anchoring instantiate each candidate in the robot frame. Multi-view observation verification converts generative candidates into trusted 3D object models. The model library and state log provide a unified interface for grasp initialization, collision checking, model reuse, and short-term continuous-frame state maintenance.

\subsection{Pose-Scale Anchoring Without Target CAD Models}

In the object-centric pipeline above, the single-view voxel generator provides only candidate geometry. Before this geometry can be used for multi-view fusion and robotic manipulation, its pose and scale relative to the real observed object must be estimated. This differs from conventional pose estimation with known CAD models. Here, the initial object model is generated by a zero-shot model and has three coupled uncertainties with respect to the real target: its canonical orientation may not match the observed orientation, its physical scale is unconstrained, and the RGB-D point cloud usually covers only the visible surface from the current view. OV-PEAR addresses these uncertainties with a two-stage pose-scale anchoring method. It first retrieves an appearance-aware initial rotation from rendered templates, and then jointly refines rotation, translation, and scale through explicit scale search and visibility-sensitive local registration.

For object $k$ observed in view $i$ with object-centric observation $(I_{k,i},B_{k,i},P_{k,i},T_i^c)$, the anchoring module outputs the similarity transformation

\[
\Theta_{k,i}=(R_{k,i},t_{k,i},s_{k,i}),
\]

where $R_{k,i}\in SO(3)$, $t_{k,i}\in\mathbb{R}^{3}$, and $s_{k,i}\in\mathbb{R}_{+}$. This transformation maps the object-canonical coordinate frame to the robot base frame or the world frame, and is used for object-canonical alignment, trusted-voxel fusion, and model-library state maintenance.

\subsubsection{Retrieval-Based Coarse Initialization}

The coarse-initialization stage does not aim to output the final pose. Its purpose is to provide a rotation initialization from which local geometric registration can converge. Since the reference model $M_{k,r}$ is generated from a single view, its canonical orientation may not be consistent with the orientation of the real object in the observation. Directly applying ICP from an arbitrary initial rotation can therefore lead to an erroneous local alignment. OV-PEAR instead builds a multi-view template feature library from the reference model and converts coarse pose search in continuous rotation space into appearance retrieval over a finite set of candidate views.

Fibonacci sphere sampling is first used to generate $K$ approximately uniform candidate viewpoints on the unit sphere:

\[
Q=\left\{q_l\right\}_{l=1}^{K}.
\]

For each candidate viewpoint $q_l$, the system renders the reference model $M_{k,r}$ to obtain a synthetic view image $I_l^{\mathrm{syn}}$, and extracts a global appearance feature using the visual encoder $\Phi(\cdot)$:

\[
f_l=\Phi(I_l^{\mathrm{syn}}),\qquad l=1,\ldots,K.
\]

This produces the template feature library

\[
\mathcal{T}_k=\left\{(f_l,R_l)\right\}_{l=1}^{K},
\]

where $R_l$ is the initial rotation matrix corresponding to candidate viewpoint $q_l$.

For the local image $I_{k,i}$ and mask $B_{k,i}$ of object $k$ in the real observation view $i$, the observed feature is

\[
f_{k,i}=\Phi(I_{k,i}\odot B_{k,i}).
\]

The cosine similarity between the observed feature and each template feature is then computed as

\[
\operatorname{Sim}(f_{k,i},f_l)
=\frac{f_{k,i}^{\top}f_l}{\lVert f_{k,i}\rVert_2\,\lVert f_l\rVert_2}.
\]

The template with the highest similarity determines the initial rotation:

\[
l^\ast=\arg\max_{l}\operatorname{Sim}(f_{k,i},f_l),
\qquad R_{k,i}^{0}=R_{l^\ast}.
\]

This step turns the canonical-view ambiguity of the generated model into a finite template retrieval problem. The subsequent local registration therefore starts from a plausible rotation neighborhood, rather than from an unconstrained search over a large rotation space.

For translation initialization, OV-PEAR uses the target local point cloud associated with the instance mask, $P_{k,i}=\{p_j\}_{j=1}^{N_i}$. Its geometric centroid is used as the initial translation:

\[
t_{k,i}^{0}=\frac{1}{N_i}\sum_{j=1}^{N_i}p_j.
\]

This gives the initial pose $(R_{k,i}^{0},t_{k,i}^{0})$ for local registration.

\subsubsection{Pose-Scale Anchoring With Template Retrieval and Visibility-Sensitive Registration}

After obtaining the initial rotation and translation, OV-PEAR performs scale-adaptive local registration. This stage addresses two issues. First, the single-view generated model has no true metric scale, so the scale factor must be explicitly estimated. Second, the observed point cloud $P_{k,i}$ contains only the visible surface under the current viewpoint, whereas the initial model $M_{k,r}$ is a complete-object hypothesis. Direct standard ICP would incorrectly force invisible model regions to participate in nearest-neighbor matching.

The system first samples a source point cloud from the surface of the reference model $M_{k,r}$:

\[
P_s=\left\{(x_m,n_m)\right\}_{m=1}^{M},
\]

where $x_m\in\mathbb{R}^{3}$ is a sampled model point, $n_m$ is its outward unit normal, and $M$ is the number of sampled points. To compensate for scale differences between the generated model and the real object, OV-PEAR constructs a candidate scale set over $[s_{\min},s_{\max}]$ with step size $\Delta s$:

\[
\mathcal{S}
=\left\{s_q\mid s_q=s_{\min}+q\Delta s,\ q=0,1,\ldots,K_s,\ s_q\le s_{\max}\right\}.
\]

The scale search range is determined by the typical scale error of the generative model on the development set and by the tolerance of the target operation. If the characteristic size of the target object is $D$ and the allowable scale error is $e_s$, the scale step should satisfy

\[
\Delta s\cdot D\le e_s,
\]

so that the geometric difference between neighboring scale candidates does not exceed the manipulation tolerance.

To reduce interference from invisible model surfaces, OV-PEAR recomputes the visibility weight during local registration instead of fixing it at the initial pose. The pose-scale state is estimated through joint pose-scale anchoring: scale is searched over a finite candidate set, and for each scale candidate, rotation and translation are refined by a finite visibility-weighted ICP process. For each $s_q\in\mathcal{S}$, visibility-weighted ICP is initialized with $(R_{k,i}^{0},t_{k,i}^{0})$. At iteration $\ell$, each source point and normal are transformed as

\[
p_m^{(\ell)}=s_qR^{(\ell)}x_m+t^{(\ell)},\qquad
n_m^{(\ell)}=R^{(\ell)}n_m .
\]

Let $o_i$ be the camera center of view $i$. The viewing direction and visibility weight are

\[
v_m^{(\ell)}
=\frac{o_i-p_m^{(\ell)}}{\left\lVert o_i-p_m^{(\ell)}\right\rVert_2},
\qquad
w_m^{(\ell)}
=\max\!\left(0,(n_m^{(\ell)})^\top v_m^{(\ell)}\right).
\]

For $p_m^{(\ell)}$, the nearest neighbor $y_m^{(\ell)}$ is searched in $P_{k,i}$, and pairs with $\|y_m^{(\ell)}-p_m^{(\ell)}\|_2\ge\tau_{\mathrm{corr}}$ are removed from the valid set $\mathcal{C}^{(\ell)}$. The pose update for the current scale candidate minimizes

\[
\begin{split}
E^{(\ell)}(R,t\mid s_q)
=
\sum_{m\in\mathcal{C}^{(\ell)}}
w_m^{(\ell)}\cdot
\left\lVert
y_m^{(\ell)}-(s_qRx_m+t)
\right\rVert_2^2 .
\end{split}
\]

The ICP refinement stops when the fixed convergence criteria of the ICP implementation are satisfied or when the maximum iteration number is reached. The locally refined pose returned for scale $s_q$ is denoted as $(R_{k,i}^{(s_q)},t_{k,i}^{(s_q)})$. Because this refinement is repeated for all $s_q$, the final anchoring result depends jointly on scale, rotation, and translation rather than on three independent estimates.

After ICP convergence, the registration quality of each scale candidate is evaluated using a visible-region fitness score. Let $\mathcal{M}_{v}^{(s_q)}$ be the set of final model points that have positive visibility weights and valid correspondences after refinement, $N_{v}^{(s_q)}=|\mathcal{M}_{v}^{(s_q)}|$, and $r_m^{(s_q)}$ be the final residual for point $m$. The fitness of $s_q$ is defined as

\[
\operatorname{Fit}(s_q)
=
\frac{1}{N_{v}^{(s_q)}}
\sum_{m\in\mathcal{M}_{v}^{(s_q)}}
\mathbb{I}\!\left(r_m^{(s_q)}<\tau_{\mathrm{fit}}\right),
\]

where $\tau_{\mathrm{fit}}$ is used only for scale selection and is distinct from the correspondence gating threshold $\tau_{\mathrm{corr}}$ and the ICP convergence thresholds.

The scale with the highest fitness is selected as the optimal scale estimate:

\[
s_{k,i}^{\ast}
=\arg\max_{s_q\in\mathcal{S}}
\operatorname{Fit}(s_q),
\]

and the corresponding rotation and translation are used as the final anchoring result:

\[
R_{k,i}=R_{k,i}^{(s_{k,i}^{\ast})},\qquad
t_{k,i}=t_{k,i}^{(s_{k,i}^{\ast})},\qquad
s_{k,i}=s_{k,i}^{\ast}.
\]

The pose-scale state of object $k$ in observation view $i$ is therefore

\[
\Theta_{k,i}=(R_{k,i},t_{k,i},s_{k,i}).
\]

This two-stage design decomposes CAD-free pose-scale estimation into appearance retrieval initialization and scale-conditioned visibility-sensitive geometric registration. The first stage reduces the canonical-orientation ambiguity of the generated model and narrows the rotation search range. The second stage compensates for scale bias while suppressing the influence of invisible surfaces and hallucinated structures under partial observations. The key is not template retrieval or local ICP alone, but the integration of appearance discriminability, scale uncertainty, and partial visibility constraints into a single anchoring process. The resulting $\Theta_{k,i}$ provides the basis for object-canonical alignment, trusted-voxel fusion, model-library matching, and continuous-frame state maintenance.

\subsection{Observation-Verified Multi-View Object Modeling}

After pose-scale anchoring, OV-PEAR has obtained the transformation parameters of object $k$ in all valid observation views:

\[
\Theta_{k,i}=(R_{k,i},t_{k,i},s_{k,i}),\qquad i\in\mathcal{I}_k.
\]

The goal of this subsection is therefore not to re-estimate object pose. Instead, OV-PEAR uses $\Theta_{k,i}$ to express multi-view generated voxel hypotheses and real RGB-D geometric evidence in the object-canonical coordinate frame, and then selects trusted object geometry. In contrast to directly using a single-view generated result, generated voxels are treated as high-recall but unverified candidate geometry. In contrast to intersection-based or purely subtractive fusion, OV-PEAR does not delete inconsistent regions at an early stage. It follows a union-first then carve strategy: it first preserves potentially correct structures from multi-view generated hypotheses, and then verifies and carves them using real depth hits, free-space ray carving~\cite{kutulakos2000theory}, and multi-view support. This design suppresses hallucinated structures in the generated 3D model while avoiding irreversible geometric deletion caused by one low-quality view, occlusion, or mask incompleteness.

\subsubsection{Union-First Candidate Voxel Aggregation}

Let the voxel hypothesis generated for object $k$ in valid view $i$ be

\[
V_{k,i}=\left\{(x_{k,i,j},\rho_{k,i,j})\right\}_{j=1}^{N_{k,i}^{v}},
\]

where $x_{k,i,j}\in\mathbb{R}^{3}$ is the voxel-center coordinate, $\rho_{k,i,j}\in[0,1]$ is the corresponding occupancy response, and $N_{k,i}^{v}$ is the number of voxels in that view.

The generated voxels $V_{k,i}$ and the real depth point cloud $P_{k,i}$ originate from different coordinate systems. We therefore use $W_{\mathrm{vox}}$ and $W_{\mathrm{dep}}$ to denote generated-voxel normalization and depth-evidence inverse transformation, respectively. Both are associated with the same pose-scale anchoring result $\Theta_{k,i}$, but their mapping directions are determined by their original coordinate systems. Specifically, $\Theta_{k,i}$ maps the object-canonical coordinate frame to the robot base frame or world frame. Since the real depth point cloud $P_{k,i}$ is already expressed in the robot base frame or world frame through the camera extrinsic parameters, it is mapped back to the object-canonical frame using $\Theta_{k,i}^{-1}$. The generated voxels $V_{k,i}$ originate from the candidate voxel space of the single-view generator, and $W_{\mathrm{vox}}$ normalizes and aligns the voxel hypothesis of that view to the shared object-canonical frame:

\[
\bar V_{k,i}=W_{\mathrm{vox}}(V_{k,i};\Theta_{k,i}).
\]

At the same time, the real depth point cloud and camera center are compensated into the same object coordinate frame:

\[
\bar P_{k,i}=W_{\mathrm{dep}}(P_{k,i};\Theta_{k,i}^{-1}),\qquad
\bar o_{k,i}=W_{\mathrm{dep}}(o_i;\Theta_{k,i}^{-1}).
\]

Here, $o_i$ is the camera center of view $i$. After this transformation, local geometric evidence collected from different camera positions, different observation times, or even rigid target motion is represented in the object's own coordinate frame.

Let the regular voxel grid in the object-canonical coordinate frame be

\[
\Omega_k=\left\{v_n\right\}_{n=1}^{N_g},
\]

where $v_n$ denotes the $n$th voxel cell. The voxel resampling operator $\Gamma(\cdot)$ projects the aligned generated voxel hypothesis onto this regular grid:

\[
\pi_{k,i}(v_n)=\Gamma(v_n,\bar V_{k,i}),\qquad
\pi_{k,i}(v_n)\in[0,1].
\]

Here, $\pi_{k,i}(v_n)$ denotes the occupancy support of the generated voxel hypothesis from view $i$ for grid voxel $v_n$.

OV-PEAR first constructs a candidate voxel union:

\[
V_{k,\mathrm{cand}}
=\left\{v_n\in\Omega_k\mid \exists i\in\mathcal{I}_k,\ \pi_{k,i}(v_n)>0\right\}.
\]

This union-first design avoids the false-negative sensitivity of intersection-based fusion. If one view produces an undersized voxel hypothesis due to occlusion, mask fragmentation, scale bias, or a single-view generation failure, intersection-based fusion would delete structures that are correctly present in other views at an early stage. Union-first fusion instead retains candidate geometry proposed by all valid views and leaves over-expanded regions to be carved by real RGB-D evidence, preserving high structural recall.

The candidate set $V_{k,\mathrm{cand}}$ is not used directly as the final object geometry. Union-first aggregation increases structural recall, but it can also introduce more potential false-positive voxels. OV-PEAR therefore treats it only as a set of candidates to be verified. Subsequent confidence updating requires each candidate voxel to obtain depth hits, multi-view generative support, or at least no conflict with visible free space, thereby preventing single-view hallucinations from being directly written into the final model.

\subsubsection{Initial Confidence Fusion Based on Visible Geometric Coverage}

The candidate voxel set $V_{k,\mathrm{cand}}$ captures structural recall from multi-view generated hypotheses, but it is not yet trusted geometry. Different views cover different visible parts of the object. If all views are accumulated with equal weight, a severely occluded view or a view containing only a small visible surface may exert excessive influence. OV-PEAR therefore assigns an initial fusion weight to each view according to the visible geometric coverage of its real observed point cloud. This weight is only an approximate measure of geometric coverage, not a complete evaluation of observation quality or sensor reliability. The actual occupancy confidence is further corrected by depth-hit and free-space evidence.

For the object-canonical point cloud of view $i$,

\[
\bar P_{k,i}=\left\{\bar p_{k,i,m}\right\}_{m=1}^{N_i},
\]

the point-cloud centroid is

\[
\mu_{k,i}=\frac{1}{N_i}\sum_{m=1}^{N_i}\bar p_{k,i,m}.
\]

The covariance matrix is

\[
C_{k,i}
=\frac{1}{N_i}\sum_{m=1}^{N_i}
(\bar p_{k,i,m}-\mu_{k,i})(\bar p_{k,i,m}-\mu_{k,i})^{\top}.
\]

Eigenvalue decomposition of $C_{k,i}$ gives three orthogonal principal directions $e_{k,i,j}$. Projecting the point cloud onto each principal direction gives the coverage range

\[
\begin{aligned}
\Delta l_{k,i,j}
&=\max_m(e_{k,i,j}^{\top}\bar p_{k,i,m})
-\min_m(e_{k,i,j}^{\top}\bar p_{k,i,m}),\\
&\qquad j=1,2,3.
\end{aligned}
\]

The visible coverage of view $i$ is then

\[
S_{k,i}=\prod_{j=1}^{3}\max(\Delta l_{k,i,j},\epsilon),
\]

where $\epsilon$ is a lower-bound threshold that prevents degeneracy when the point cloud has a very small range along one direction. The corresponding view-fusion weight is

\[
\alpha_{k,i}=\frac{S_{k,i}}{\sum_{q\in\mathcal{I}_k}S_{k,q}}.
\]

The initial confidence of a candidate voxel is defined as

\[
C_0(v_n)=\sum_{i\in\mathcal{I}_k}\alpha_{k,i}\pi_{k,i}(v_n),
\qquad v_n\in V_{k,\mathrm{cand}}.
\]

In this definition, multi-view support is not introduced as an additional discrete rule. It is accumulated in $C_0(v_n)$ through the occupancy responses $\pi_{k,i}(v_n)$ and the coverage weights $\alpha_{k,i}$ of different views. Voxels jointly predicted by multiple valid views receive higher initial confidence, whereas responses accidentally produced by a single low-coverage view are more likely to be suppressed by later thresholding or free-space carving. Thus, $C_0(v_n)$ represents multi-view support at the generative-prior level, but it still requires verification by real RGB-D geometric evidence.

\subsubsection{Trusted Voxel Selection Using Depth Hits and Ray Carving}

To improve the efficiency of nearest-neighbor queries between voxel centers and real depth surfaces, OV-PEAR builds a KD-tree for each object-canonical point cloud $\bar P_{k,i}$ and uses it to determine whether a candidate voxel is supported by the visible depth surface. Free-space judgment, however, does not rely only on KD-tree nearest-neighbor distance. It is determined by depth ordering. For a candidate voxel $v_n$, the system projects its center $c_n$ into the depth map of view $i$ through the current object transformation and camera extrinsic parameters, obtaining the corresponding pixel location and projected voxel depth. If the voxel lies in front of the observed depth surface and no local real-surface hit is found in its neighborhood, it is classified as free space. If it agrees with the observed depth surface within $\epsilon_h$, it is classified as a hit. If it lies behind the observed surface or the current view provides no valid depth, it is classified as occluded or unknown.

Formally, for $v_n\in V_{k,\mathrm{cand}}$, let $c_n$ be its center in the object-canonical coordinate frame. In view $i$, it is transformed to the camera coordinate frame and projected as

\[
(u_{k,i,n},z_{k,i,n})
=\Pi_i\!\left((T_i^c)^{-1}\Theta_{k,i}\tilde c_n\right),
\]

where $\tilde c_n$ is the homogeneous coordinate of $c_n$, $u_{k,i,n}$ is the projected pixel location, $z_{k,i,n}$ is the projected depth of the voxel center in that view, and $\Pi_i(\cdot)$ is the camera projection model. Let the observed depth at that pixel be $D_i(u_{k,i,n})$. The evidence state is defined as follows:

if $|z_{k,i,n}-D_i(u_{k,i,n})|<\epsilon_h$, then $v_n$ is classified as a depth hit;

if $z_{k,i,n}<D_i(u_{k,i,n})-\epsilon_h$ and the KD-tree query does not find a local real surface, then $v_n$ is classified as free space;

if $z_{k,i,n}>D_i(u_{k,i,n})+\epsilon_h$, or if the pixel depth is invalid, then $v_n$ is treated as occluded or unknown.

The threshold $\epsilon_h$ is set at a scale consistent with voxel side length, depth noise, and calibration error, so that isolated noise or boundary quantization does not cause erroneous hit decisions.

To avoid order dependence from sequential per-view updates, OV-PEAR first counts the geometric evidence obtained by each candidate voxel across all views:

\[
\begin{aligned}
h_n&=\sum_{i\in\mathcal{I}_k}
\mathbb{I}(v_n\ \text{is hit in view}\ i),\\
f_n&=\sum_{i\in\mathcal{I}_k}
\mathbb{I}(v_n\ \text{is free space in view}\ i),\\
g_n&=\sum_{i\in\mathcal{I}_k}
\mathbb{I}\!\left(\pi_{k,i}(v_n)>0\right).
\end{aligned}
\]

Here, $h_n$ and $f_n$ denote the number of views that classify the voxel as a hit and as free space, respectively. The value $g_n$ denotes the number of generated-view hypotheses that support the voxel.

Based on this evidence, the updated voxel confidence is defined as

\begin{equation}
C(v_n)=
\begin{cases}
\max\!\left(C_0(v_n),p_{\mathrm{hit}}\right), & h_n>0,\\
\min\!\left(C_0(v_n),p_{\mathrm{free}}\right), & h_n=0\ \mathrm{and}\ f_n>0,\\
C_0(v_n), & h_n=0\ \mathrm{and}\ f_n=0.
\end{cases}
\label{eq:voxel-confidence-update}
\end{equation}

Here, $p_{\mathrm{hit}}=0.95$ is the high-confidence lower bound for depth-hit voxels, and $p_{\mathrm{free}}=0.05$ is the low-confidence upper bound for free-space voxels. They can be interpreted as bounded probabilistic relaxations of hard 1/0 decisions. In controlled industrial RGB-D observations, hits and free-space evidence provide strong support and strong contradiction, respectively, while still preserving uncertainty caused by depth noise, segmentation error, and hand-eye calibration error.

When the same voxel receives both hit and free-space evidence, OV-PEAR gives priority to hit evidence. A depth-surface hit provides direct local occupancy support, whereas free-space classification is more sensitive to registration error, depth-boundary quantization, mask-boundary error, and hand-eye calibration residuals. This hit-prioritized update is a conservative strategy that reduces the risk of over-carving real thin structures, local boundaries, or small protrusions. For candidate voxels that lack hit support and are stably classified as conflicting with observed free space, $p_{\mathrm{free}}$ suppresses their confidence.

The final trusted voxel set is

\[
\begin{aligned}
V_{k,\mathrm{trust}}
&=\left\{v_n\in V_{k,\mathrm{cand}}\,\middle|\,
C(v_n)\ge\tau_c,\right.\\
&\qquad\left.
h_n\ge\tau_h\ \mathrm{or}\ (h_n=0\ \mathrm{and}\ g_n\ge\tau_s)
\right\}.
\end{aligned}
\]

Here, $\tau_c$ is the global confidence threshold, $\tau_h$ is the minimum number of real depth hits, and $\tau_s$ is the minimum multi-view generative support required when no depth hit is available. A voxel predicted only accidentally by a single-view prior and lacking depth-hit or multi-view support is unlikely to pass the threshold. In contrast, real object structures usually receive depth hits or consistent cross-view support and are therefore retained.

A voxel that is not directly hit by the current view is not deleted immediately. If it receives multi-view generative support and does not conflict with visible free space, it can be retained as an occluded or incompletely observed structure. Conversely, voxels that clearly conflict with visible depth and lack hit or multi-view support are suppressed.

OV-PEAR therefore implements a union-first then carve observation-verification mechanism. Union-first aggregation maintains structural recall from generated candidate geometry. Depth hits provide positive surface evidence. Free-space ray carving provides negative spatial evidence. Occluded or unknown regions are not forcibly removed by a single view. This mechanism allows the final voxel set to use generative priors to complement unobserved structures while suppressing hallucinated regions and local errors using real RGB-D observations.

\subsubsection{Trusted-Voxel and RGB/Mask-Guided Mesh Decoding}

After obtaining $V_{k,\mathrm{trust}}$, OV-PEAR combines the reference RGB/mask information to generate the final object mesh:

\[
M_k^\ast=D_{\mathrm{mesh}}(V_{k,\mathrm{trust}},I_{k,r},B_{k,r}).
\]

Here, $V_{k,\mathrm{trust}}$ provides trusted geometric support in the object-canonical coordinate frame, $I_{k,r}$ and $B_{k,r}$ provide appearance, boundary, and color information, and $D_{\mathrm{mesh}}(\cdot)$ denotes a pretrained mesh decoder that takes voxel occupancy and RGB/mask appearance as input to produce a colored object mesh. The decoder itself is not claimed as the main methodological contribution. Instead, OV-PEAR replaces its geometric input from a single-view voxel prediction with $V_{k,\mathrm{trust}}$, which has been verified by multi-view RGB-D evidence, thereby constraining the occupiable region, scale range, and structural boundary of the final model.

Compared with directly decoding any single-view generated voxel grid, mesh generation based on $V_{k,\mathrm{trust}}$ reduces local structural errors caused by single-view occlusion and generative-prior bias. Compared with extracting a mesh only from the voxel surface, incorporating RGB/mask information helps produce a more continuous object surface and a more stable appearance representation.

The final object model $M_k^\ast$ is inserted into the object-model library:

\[
L\leftarrow L\cup\{M_k^\ast\}.
\]

This model replaces the original single-view generated model as the object-level geometric template for subsequent pose estimation, model-library matching, grasp initialization, and collision checking. OV-PEAR thereby unifies single-view generative candidates, multi-view pose-scale anchoring, and real RGB-D observation verification, completing the conversion from a generative object hypothesis to an observation-verified 3D object model. After insertion into the library, $M_k^\ast$ also serves as the shared geometric template for continuous-frame state maintenance and object-level scene replay; its reuse and compression effects are evaluated in the experiments.

\section{Experiment}

This section evaluates OV-PEAR along four axes: pose-estimation accuracy, object-model quality, robotic manipulation initialization, and short-term object-state maintenance. The experiments are organized as public industrial benchmarking, real workbench perception, object-model reuse, connector assembly, and ablation studies. Implementation parameters are fixed across objects and scenes; the full parameter list, runtime accounting, USB tolerance calibration, and auxiliary visualizations are provided in Appendices~\ref{app:parameters-runtime}--\ref{app:state-maintenance-visualizations}.

\subsection{Experimental Setup}

The robotic platform consists of a Franka Panda manipulator, a Robotiq gripper, and a wrist-mounted Intel RealSense RGB-D camera in an eye-in-hand configuration. During object modeling, the robot follows pre-defined observation trajectories, captures multi-view RGB-D data, and uses calibrated camera extrinsics to express all observations in a common robot coordinate frame. Unless otherwise stated, the target object is static within one multi-view modeling window. The short-term state-maintenance experiment uses a different mode: after a verified object model is inserted into the model library, subsequent frames reuse the model and update only pose, scale, confidence, identity, and visibility states.

For public industrial benchmarking, we use the images, depth maps, annotations, and evaluation scripts provided by BOP-Industrial and the BOP evaluation ecosystem~\cite{hodan2018bop,hodan2024bopchallenge2023}. For the real workbench and connector experiments, the robot actively acquires multi-view RGB-D observations. The real workbench evaluation set contains 15 scenes, four calibrated views per scene, 352 scene-camera records, 2264 per-view pose/visibility annotations, and 10 object models. All experiments are executed on a workstation equipped with an NVIDIA RTX 4090D GPU.

We compare OV-PEAR with FRTPose-WAPR, FoundationPose, and SAM-6D. FRTPose-WAPR and FoundationPose~\cite{wen2024foundationpose} represent competitive industrial and model-based novel-object pose-estimation pipelines, while SAM-6D~\cite{lin2024sam6d} represents recent zero-shot 6D pose estimation based on visual foundation models. On BOP-Industrial, all methods are evaluated under the same official single-view AR protocol. In real workbench scenes, object detection and segmentation are evaluated by mAP@0.5:0.95, pose accuracy by object-center error and frame-aligned rotation error, and geometry by voxel IoU. For short-term state maintenance, we evaluate object-state error, replay quality, and representation efficiency. For connector assembly, we report grasp success, insertion success, and assembly success.

\paragraph{Metric definitions}
The reported accuracy metrics follow the protocol of each experiment and are not cross-compared across protocols. In the public BOP-Industrial benchmark, we follow the official single-view BOP evaluation setting and report AR for pose estimation. In the real workbench perception experiment, mAP@0.5:0.95 evaluates instance detection and segmentation quality, while object-center error, frame-aligned rotation error, and voxel IoU evaluate pose-scale estimation and geometric modeling. In the core perception ablation, AR is reported only for the corresponding 6D detection protocol, whereas voxel IoU measures the quality of the verified object geometry.

For the real workbench benchmark, dataset-provided object masks are used to define object crops and visible object regions for pose evaluation, so that the comparison focuses on pose-scale estimation and geometric modeling rather than mask quality. In subsequent online manipulation and state-maintenance experiments, object masks are generated by SAM3. Object-center error is computed as the Euclidean distance between transformed 3D bounding-box centers in the robot/world frame, rather than as the distance between arbitrary model origins.

Because the reconstructed OV-PEAR mesh has an object-canonical frame that is not guaranteed to match the coordinate frame of the reference object model, rotation error is evaluated after a fixed object-level frame correction. Let \(C\) denote the reconstructed mesh frame, \(M\) the reference model frame, and \(A_{MC}\) the fixed transform from \(C\) to \(M\), estimated once per object from the final reconstructed mesh and the reference model. For each evaluation frame, the corrected ground-truth pose of the reconstructed mesh is
\[
T^{gt}_{WC}=T^{gt}_{WM}A_{MC}.
\]
The reported rotation error is computed between \(T^{pred}_{WC}\) and \(T^{gt}_{WC}\). This correction is fixed at the object level and is not fitted per frame. For symmetric or near-symmetric objects, equivalent rotations are handled by taking the minimum error over the valid symmetry set when such symmetry is geometrically supported.

\subsection{Comparative Experiments}

\subsubsection{Public Industrial Benchmark}

\begin{figure*}[!t]
	\centering
	\includegraphics[width=\textwidth]{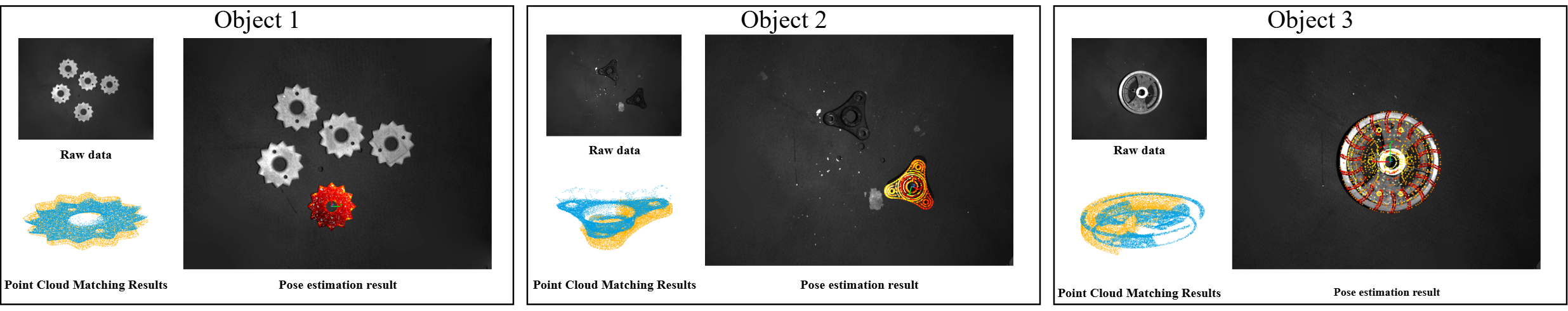}
	\caption{Qualitative object-instantiation and pose-scale estimation results on representative BOP-Industrial objects. For each object, the input observation is shown on the left, the middle visualization shows the alignment between the generated object model and the observed point cloud, and the right visualization overlays the reconstructed object model in the scene using the estimated 6D pose and scale.}
	\label{fig:bop-qualitative}
\end{figure*}

\begin{table*}[!t]
	\centering
	\scriptsize
	\caption{Pose-estimation results on three BOP industrial datasets using AR as the evaluation metric}
	\label{tab:source-2}
	\begin{adjustbox}{max width=\textwidth}
		\begin{tabular}{lccccc}
			\toprule
			Method & Views & IPD & XYZ-IBD & ITODD-MV & Average \\
			\midrule
			FRTPose-WAPR & 1 & 0.481 & 0.516 & 0.688 & 0.562 \\
			FoundationPose & 1 & 0.408 & 0.564 & 0.512 & 0.495 \\
			SAM-6D & 1 & 0.415 & 0.578 & 0.595 & 0.529 \\
			OV-PEAR & 1 & 0.609 & 0.647 & 0.715 & 0.657 \\
			\bottomrule
		\end{tabular}
	\end{adjustbox}
	\vspace{0.3em}
	\begin{minipage}{0.98\textwidth}
		\footnotesize Note: FRTPose-WAPR, FoundationPose, and SAM-6D were reproduced using their official GitHub repositories or public implementations and evaluated under the same BOP-Industrial single-view input setting, data subsets, and BOP official AR scripts. The corresponding commits, configuration files, and execution scripts will be listed in the reproducibility package.
	\end{minipage}
\end{table*}
BOP-Industrial contains IPD, XYZ-IBD, and ITODD-MV subsets collected in real industrial environments with textureless, reflective, symmetric, densely stacked, and occluded objects~\cite{kleeberger2019ipd,drost2017itodd}. To keep this comparison aligned with the official benchmark setting, OV-PEAR is evaluated in a single-view mode and only pose-estimation AR is reported. Multi-view voxel optimization, robot manipulation, and continuous-frame replay are not mixed into this comparison.

As shown in Table~\ref{tab:source-2}, OV-PEAR obtains the highest AR on all three subsets and reaches an average AR of 0.657, improving over FRTPose-WAPR, FoundationPose, and SAM-6D by 0.095, 0.162, and 0.128, respectively. Relative to the strongest baseline on each subset, OV-PEAR improves AR by 0.128 on IPD, 0.069 on XYZ-IBD, and 0.027 on ITODD-MV, with a mean best-baseline gain of 0.075 and a sample variance of 0.00257 across the three subsets. Relative to the three-baseline mean after excluding OV-PEAR, the gains are 0.174, 0.094, and 0.117 AR on IPD, XYZ-IBD, and ITODD-MV, respectively. Because this experiment uses only the official single-view benchmark inputs, the gain cannot be attributed to multi-view voxel fusion or downstream manipulation feedback. It mainly reflects the effect of pose-scale anchoring: the generated object hypothesis is not accepted as a free-form prior, but is explicitly scaled and registered against the observed RGB-D evidence.

The subset-wise pattern is also informative. The improvement is largest on IPD and XYZ-IBD, where clutter, weak texture, and industrial object ambiguity make purely template-based or frame-wise zero-shot estimation unstable. On ITODD-MV, where FRTPose-WAPR is already competitive, OV-PEAR still improves AR but by a smaller margin. This pattern is directionally consistent across the three subsets, but the number of benchmark subsets is too small for a high-power significance test. A one-sample test on the three best-baseline gains gives a positive trend but should be interpreted descriptively rather than as definitive inferential evidence. The public benchmark therefore supports the front-end pose-estimation component, but it does not by itself evaluate the main reconstruction and reuse claims, which are tested in the following real-workbench and state-maintenance experiments.

\subsubsection{Unseen-Object Perception in Real Workbench Scenes}

The real workbench experiment evaluates whether OV-PEAR generalizes beyond fixed benchmark categories and layouts. Empty bottles, snack packages, glasses, and other objects are randomly placed in a cluttered container. The wrist-mounted RGB-D camera observes the scene from multiple viewpoints, and the system estimates object category, pose, scale, and geometry. Since not all baselines support segmentation, multi-view input, or object-level reconstruction, unsupported outputs are marked as N/A rather than treated as execution failures.

\begin{table*}[!t]
	\centering
	\scriptsize
	\caption{Perception performance in real workbench scenes with unseen objects.}
	\label{tab:source-3}
	\begin{adjustbox}{max width=\textwidth}
		\begin{tabular}{lcccccc}
			\toprule
			Method & mAP & Center err. (cm) & Aligned rot. (deg) & Voxel IoU & Build (s) & Pose (s) \\
			\midrule
			FRTPose-WAPR & N/A & 8.23 & 19.21 & N/A & N/A & 0.72 \\
			FoundationPose & N/A & 12.56 & 26.30 & N/A & N/A & 4.53 \\
			SAM-6D & 0.428 & 6.95 & 21.87 & N/A & N/A & 0.28 \\
			OV-PEAR-SV & 0.415 & 2.88 & 15.93 & 0.613 & 12.61 & 0.53 \\
			OV-PEAR-4V & 0.415 & 1.28 & 13.17 & 0.689 & 33.45 & 0.53 \\
			\bottomrule
		\end{tabular}
	\end{adjustbox}
	\vspace{0.3em}
	\begin{minipage}{0.98\textwidth}
		\footnotesize
		Note: mAP is evaluated at IoU thresholds from 0.5 to 0.95. Center error measures the transformed 3D bounding-box center error. Aligned rotation error is computed after a fixed object-level frame correction. Build time denotes one-time object-model construction cost, while pose time denotes per-query pose estimation cost. N/A indicates that the method does not natively provide the corresponding output.
	\end{minipage}
\end{table*}

Table~\ref{tab:source-3} shows that OV-PEAR provides stronger overall perception performance in cluttered unseen-object scenes. In detection and segmentation, OV-PEAR and SAM-6D achieve comparable mAP, which indicates that the subsequent pose and geometry improvements are not simply caused by a stronger segmentation front end. In pose estimation, OV-PEAR-SV already reduces center error to 2.88 cm, and OV-PEAR-4V further reduces it to 1.28 cm. The four-view setting also lowers frame-aligned rotation error from 15.93 deg to 13.17 deg and increases voxel IoU from 0.613 to 0.689. These changes isolate the value of observation verification: additional views do not only add more surface points, but also suppress unsupported generative geometry and improve the consistency between the object model and the robot-frame pose.

\begin{figure*}[!t]
	\centering
	\includegraphics[width=\textwidth]{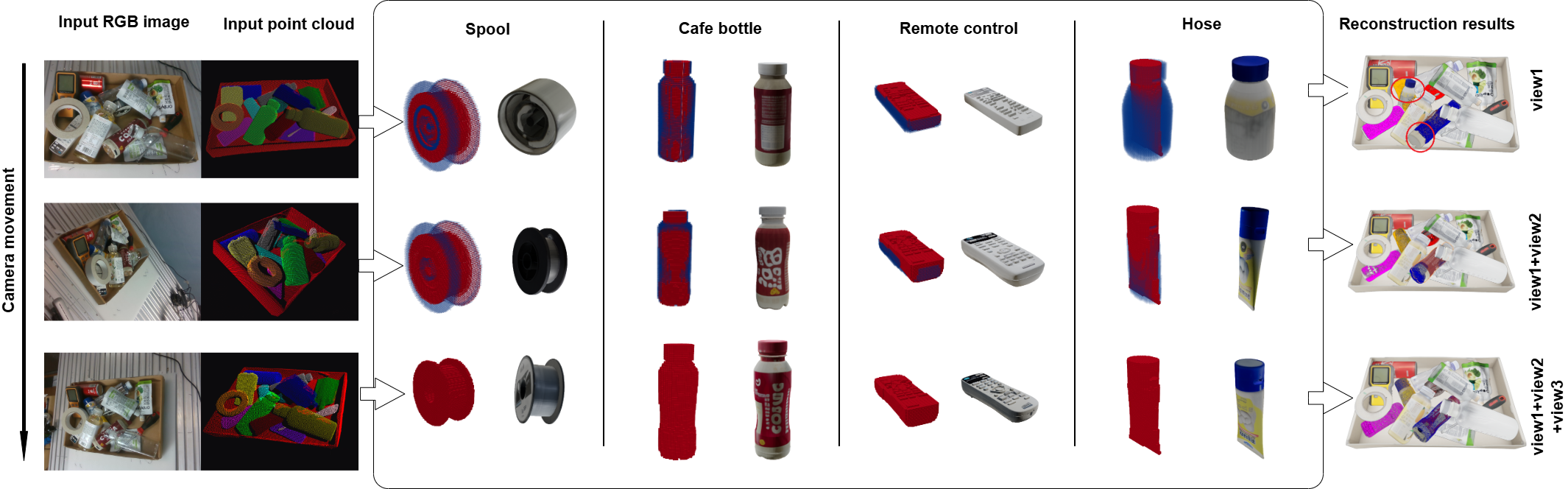}
	\caption{Qualitative results of multi-view observation verification in real workbench scenes with unseen objects. Spool, cafe bottle, remote control, and hose are shown as representative targets. Blue voxels denote hallucinated or weakly supported regions suppressed by real geometric evidence, while red voxels denote trusted regions supported by cross-view observations.}
	\label{fig:workbench-qualitative}
\end{figure*}

The remaining errors clarify the boundary of the method. Frame-aligned rotation error is reduced less strongly than center error because several workbench objects have weak texture, partial symmetry, or incomplete side observations. In such cases, multi-view verification can remove implausible occupied voxels and improve the usable object extent, but it cannot fully resolve all orientation ambiguity without more informative viewpoints or contact feedback. This is consistent with the design goal of OV-PEAR: the method is intended to produce a manipulation-ready pose-scale state and reusable object model for previously unseen objects, rather than a globally optimal dense reconstruction from arbitrary sparse views.

The initial model-building time of OV-PEAR is longer than that of pure pose-estimation methods, because this stage constructs a verified object model and writes it into the feature library. This cost is therefore paid when a new object model is introduced, not at every pose update. Once the library is available, subsequent pose-scale estimation reuses the verified object model and takes 0.53 s per instance. The update is slightly slower than SAM-6D because OV-PEAR additionally performs scale alignment and visibility-sensitive registration, but it yields a reusable object model and lower pose errors. This trade-off is appropriate for industrial workcells in which a previously unseen object may be observed once and then manipulated or localized repeatedly.

Qualitatively, Fig.~\ref{fig:workbench-qualitative} shows that OV-PEAR separates target objects in multi-object clutter and uses multi-view evidence to correct unsupported single-view geometry. The progressive removal of hallucinated voxels and the preservation of observation-supported regions explain why the multi-view model is useful for manipulation: it improves geometric plausibility, pose-scale consistency, and model reuse in the robot frame.

\subsubsection{Short-Term State Maintenance Through Object-Model Reuse}

We further evaluate whether a verified rigid-object model can be reused in a short-horizon industrial workcell. OV-PEAR first constructs an object-model library from multi-view RGB-D observations and then records only the pose, scale, confidence, visibility, and identity of each object instance in subsequent frames. The experiment does not aim to replace general dynamic SLAM or open-set scene mapping systems~\cite{millane2024nvblox,keetha2024splatam,jatavallabhula2023conceptfusion}; it tests whether object-level state can be preserved compactly after verification.

\begin{figure*}[!t]
	\centering
	\subfloat[Object-model construction and world-frame replay.]{
		\includegraphics[width=0.8\textwidth]{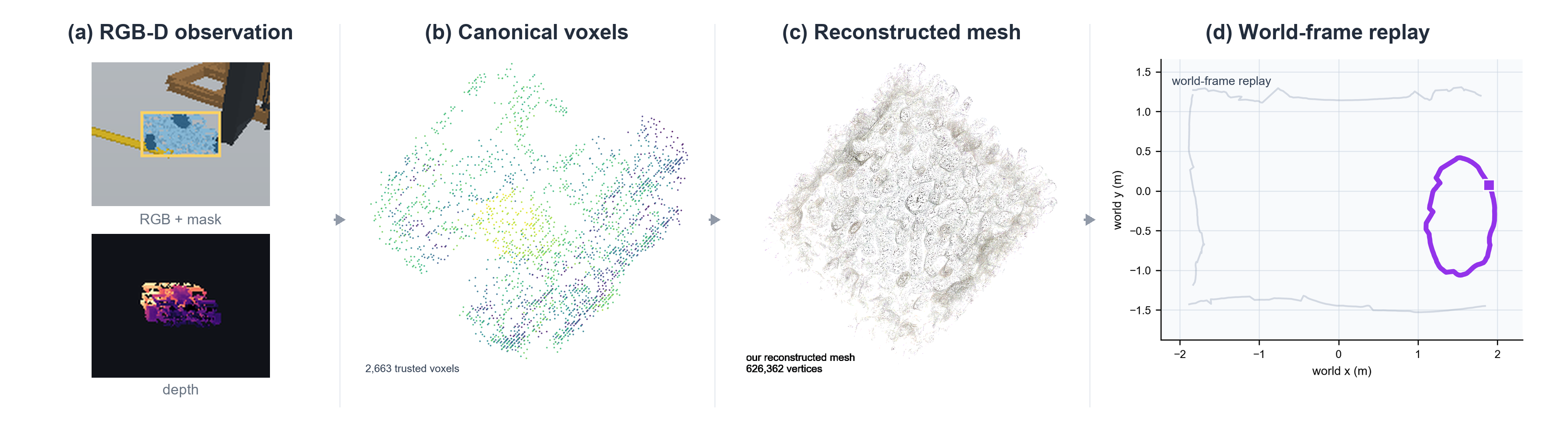}
	}
	
	\vspace{0.4em}
	
	\subfloat[Same-frame local reconstruction comparison.]{
		\includegraphics[width=0.8\textwidth]{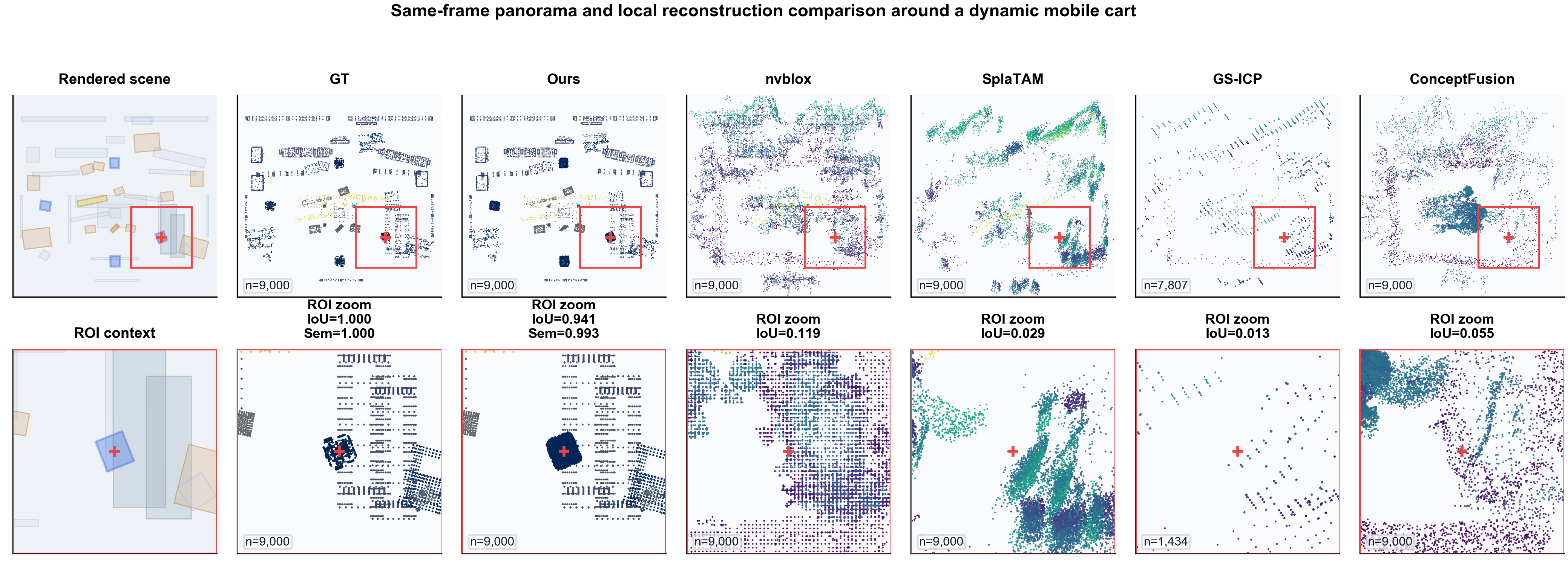}
	}
	\caption{Qualitative evidence for short-term object-state maintenance. 
		(a) OV-PEAR constructs an observation-verified object model from RGB-D evidence and reuses it through compact object-level state logging. 
		(b) Same-frame local reconstruction comparison around a dynamic mobile cart. OV-PEAR preserves a compact object-level instance, whereas map-oriented baselines mainly reconstruct visible surfaces without persistent object identity.}
	\label{fig:state-maintenance-qualitative}
\end{figure*}

\begin{table*}[!t]
\centering
\scriptsize
\caption{Audit of scene-representation capability for short-term object-state maintenance}
\label{tab:source-4}
\begin{adjustbox}{max width=\textwidth}
\begin{tabular}{lcccc}
\toprule
Method & Obj. state err. T/S & Replay Sp./Sem. & Visible F/Chamfer-L1 & Eff. Comp./Init/Upd. \\
\midrule
OV-PEAR & 0.170 / 4.36\% & 0.7800 / 0.9998 & 0.5638 / 0.1485 & $1.52\times$ / 455.06 s / 0.02 ms \\
nvblox & N/A & 0.1070 / N/A & 0.6532 / 0.0972 & $17.43\times$ / 12.31 s / 14.66 ms \\
SplaTAM & N/A & 0.0230 / N/A & 0.0423 / 0.2739 & $0.64\times$ / 134.35 s / 1119.54 ms \\
GS-ICP & N/A & 0.0074 / N/A & 0.2687 / 0.1871 & $170.80\times$ / 17.72 s / 147.69 ms \\
ConceptFusion & N/A & 0.0930 / N/A & 0.9593 / 0.0331 & $0.01\times$ / 151.76 s / 180.66 ms \\
\bottomrule
\end{tabular}
\end{adjustbox}
\vspace{0.3em}
\begin{minipage}{0.98\textwidth}
\footnotesize Note: This table audits usable outputs for short-term object-state maintenance rather than ranking general SLAM systems. Dynamic Spatial/Semantic is computed only for methods with stable object identity, object state, and replayable instance trajectories. N/A denotes an unsupported native output type.
\end{minipage}
\end{table*}

OV-PEAR obtains 0.7800 / 0.9998 for Dynamic Spatial/Semantic, with normalized object-state error of 0.170, scale error of 4.36\%, and incremental update time of 0.02 ms/frame after library construction. The near-unity semantic score indicates stable object identity and model-library indexing, whereas the spatial score is still affected by mesh-size mismatch, voxelization resolution, local occlusion, and residual state-estimation error. The result therefore supports reusable object-level state consistency rather than optimal point-level visible-surface fitting.

This distinction is important for interpreting the baseline comparison. As shown qualitatively in Fig.~\ref{fig:state-maintenance-qualitative}, nvblox and ConceptFusion provide strong visible-surface reconstructions, and Gaussian-splatting methods provide efficient differentiable scene representations~\cite{kerbl2023gaussian}; however, their native outputs do not directly encode replayable object identity, scale, and per-object state. OV-PEAR instead compresses the dynamic part of the scene into a verified object model plus a lightweight state log. It is therefore better suited to short-horizon industrial object maintenance, while general SLAM systems remain more appropriate for dense static mapping. The stage-wise runtime interpretation is given in Appendix~\ref{app:parameters-runtime}, and additional layout and ablation visualizations are given in Appendix~\ref{app:state-maintenance-visualizations}.

\subsection{Robotic Manipulation Experiments}

The connector experiments test whether the estimated pose-scale state and verified geometry can initialize real grasping and compliant insertion. USB, HAN10, and RJ45 connectors differ in insertion geometry and tolerance. To keep the main text focused on perception value, the detailed USB tolerance calibration and extended failure breakdown are moved to Appendix~\ref{app:usb-failure}; the key operating threshold is that the current compliant controller becomes unreliable when the initial pose error exceeds approximately \(5\,\mathrm{mm}\) or \(3^\circ\).

\begin{table*}[!t]
\centering
\scriptsize
\caption{Pose usability of different perception initialization methods in the USB task}
\label{tab:usb-init-comparison}
\begin{adjustbox}{max width=\textwidth}
\begin{tabularx}{\textwidth}{p{0.14\textwidth}p{0.18\textwidth}ccccX}
\toprule
Method & Prior condition & Trials & Within tolerance & Unusable & Gross outlier & Main failure cause \\
\midrule
SAM-6D & No target CAD; zero-shot pose estimation & 50 & 3/50 & 47/50 & Observed & Orientation ambiguity, center bias, and unstable frame-wise estimates \\
ICP-based & Accurate object model required & 50 & 43/50 & 7/50 & Rare & Degenerate viewpoints and weak local geometric constraints \\
OV-PEAR single construction & No target CAD; generated and verified object model & 50 & 42/50; 42/47 & 8/50 & 0/50 & Three unusable near top-down meshes; five residual errors outside tolerance \\
OV-PEAR verified-model reuse & No target CAD; verified model-library reuse & 20 & 20/20 & 0/20 & 0/20 & No evident failure \\
\bottomrule
\end{tabularx}
\end{adjustbox}
\end{table*}

Table~\ref{tab:usb-init-comparison} isolates the front-end perception contribution. We define a catastrophic pose outlier as an estimate with center error larger than \(5\,\mathrm{cm}\), a clear orientation flip, or an initialization that sends grasping or insertion away from the target region. SAM-6D produces only 3 usable poses in 50 trials because of large angular errors or center shifts. The ICP-based baseline reaches 43/50 usable trials but requires an accurate object model. OV-PEAR produces usable models in 47/50 trials, with 42 poses inside the compliant-control capture range and no catastrophic pose outlier. When the verified USB model is reused, all 20 random-initial-pose trials fall inside the tolerance range.

\paragraph{Failure Case Analysis}
The failure cases show that OV-PEAR changes the failure profile rather than simply increasing the success count. SAM-6D mainly fails through gross pose outliers, orientation ambiguity, and center bias, which place the connector outside the compliant controller's capture range. ICP is stronger when an accurate object model is supplied, but its failures arise from degenerate local geometry and weak convergence constraints. OV-PEAR failures are less catastrophic: three unusable models come from insufficient side-view geometry under near top-down observation, and five additional trials exceed the \(5\,\mathrm{mm}\) or \(3^\circ\) tolerance because of small residual pose errors. The end-to-end failures in Table~\ref{tab:connector-end-to-end} then mainly reflect post-grasp connector shift, socket-side angular error, hand-eye residuals, robot repeatability, and contact jamming rather than gross target localization errors. A fuller breakdown is provided in Appendix~\ref{app:usb-failure}.

\begin{table}[!t]
\centering
\scriptsize
\caption{Grasping and compliant-insertion results for three connector types}
\label{tab:connector-end-to-end}
\begin{adjustbox}{max width=\columnwidth}
\begin{tabular}{lccc}
\toprule
Task type & Grasp success rate & Insertion success rate & Assembly success rate \\
\midrule
USB assembly & 99/100 & 86/99 & 86\% \\
HAN10 assembly & 100/100 & 71/100 & 71\% \\
RJ45 assembly & 100/100 & 95/100 & 95\% \\
Average & 99.67\% & 84.29\% & 84.00\% \\
\bottomrule
\end{tabular}
\end{adjustbox}
\end{table}

\begin{figure*}[!t]
\centering
\includegraphics[width=\textwidth]{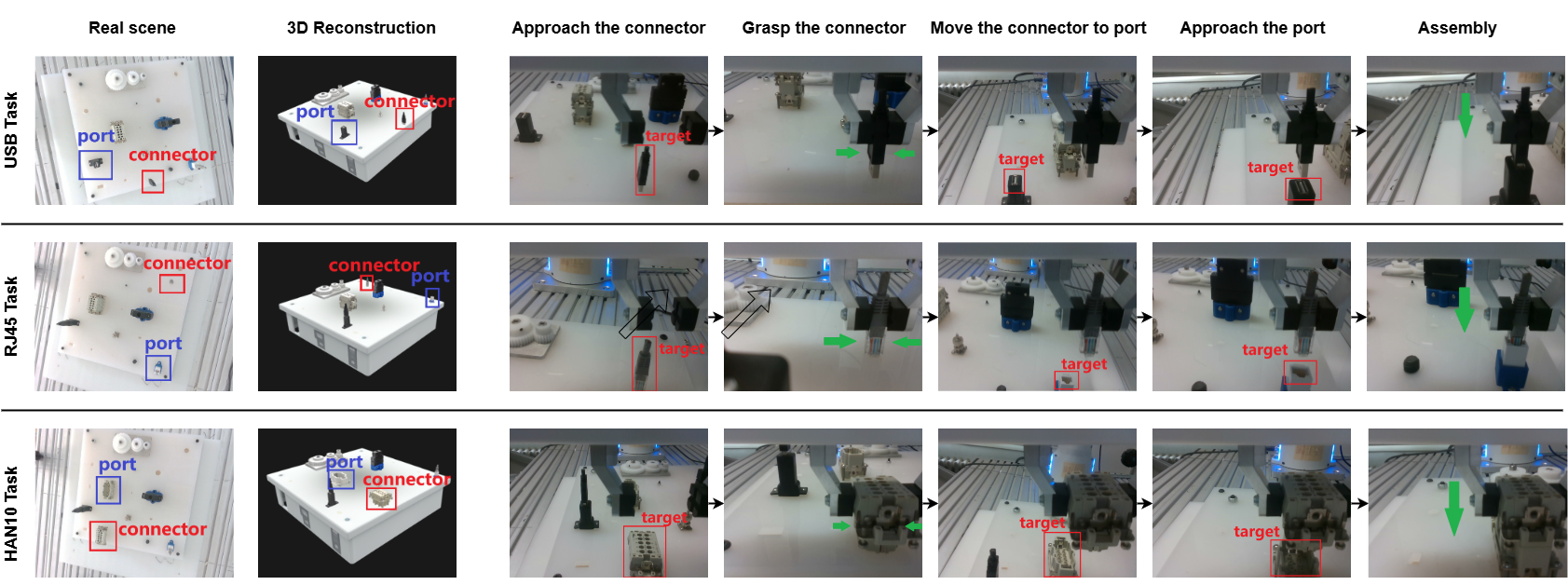}
\caption{Fine assembly processes for USB, RJ45, and HAN10 connectors. The figure shows representative procedures including multi-view perception, target grasping, pose adjustment, and compliant insertion.}
\label{fig:connector-assembly}
\end{figure*}

The end-to-end connector trials further show that OV-PEAR provides stable initialization for real manipulation. The average grasp success rate is 99.67\%, and the average assembly success rate is 84.00\%. The remaining insertion failures follow task tolerance: HAN10 is most sensitive because its multi-pin structure amplifies angular error, while RJ45 reaches 95\% assembly success. These failures mainly originate from post-grasp connector shift, socket-side angular error, hand-eye residuals, robot repeatability, and contact jamming rather than gross target localization errors. Further gains for low-tolerance connectors require socket-side perception, in-gripper pose estimation, force feedback, or visual servoing.

\subsection{Ablation Studies}

We conduct two ablation studies. The first tests whether the core perception modules are necessary for object instantiation and observation-verified modeling. The second tests whether the reuse mechanism preserves object-level state efficiently and stably.

\subsubsection{Core Perception Modules}

\begin{table*}[!t]
\centering
\scriptsize
\caption{Ablation results for core perception modules using AR as the 6D detection metric}
\label{tab:core-ablation}
\begin{adjustbox}{max width=\textwidth}
\begin{tabular}{lcccc}
\toprule
Module & Setting & Views & AR & Voxel IoU \\
\midrule
Coarse initialization & Feature-library rotation initialization & 3 & 0.762 & 0.851 \\
Coarse initialization & Without coarse initialization & 3 & 0.283 & 0.407 \\
Optimal-scale search & With parallel optimal-scale search & 3 & 0.762 & 0.851 \\
Optimal-scale search & Without optimal-scale search & 3 & 0.194 & 0.374 \\
Multi-view voxel optimization & Single view & 1 & 0.715 & 0.760 \\
Multi-view voxel optimization & Multi-view & 2 & 0.733 & 0.825 \\
Multi-view voxel optimization & Multi-view & 3 & 0.762 & 0.851 \\
\bottomrule
\end{tabular}
\end{adjustbox}
\end{table*}

Removing feature-library-based coarse initialization decreases AR from 0.762 to 0.283 and voxel IoU from 0.851 to 0.407. Without a reasonable initial rotation, the subsequent ICP refinement often converges to an incorrect local optimum, making the generated object model difficult to anchor reliably. Removing parallel optimal-scale search further reduces AR to 0.194 and voxel IoU to 0.374, showing that scale error affects not only object-size consistency but also the registration residual and subsequent geometry modeling. These results support the two-stage pose-scale alignment strategy: coarse initialization provides a convergent pose prior, and scale search compensates for size mismatch between the generated object model and real observations.

For multi-view voxel-confidence updating, the single-view setting obtains AR of 0.715 and voxel IoU of 0.760. With two views, the scores increase to 0.733 and 0.825. With three views, they reach 0.762 and 0.851. Multi-view observations therefore both recover areas occluded in a single view and suppress hallucinated structures through cross-view consistency, producing a more reliable object template. Together, coarse initialization, scale refinement, and multi-view voxel updating support OV-PEAR through pose convergence, scale consistency, and geometric trustworthiness.

\subsubsection{Object-Model Reuse and State Maintenance}

\begin{table*}[!t]
\centering
\scriptsize
\caption{Internal ablation of the object-model reuse mechanism}
\label{tab:reuse-ablation}
\begin{adjustbox}{max width=\textwidth}
\begin{tabular}{lcccc}
\toprule
Variant & State err. T/Y/S & Replay Sp./ID & Coverage & Eff. Comp./Upd./Cost \\
\midrule
Full & 0.170 / \(10.55^\circ\) / 4.36\% & 0.779 / 0.992 & 100.0\% & $1.52\times$ / 0.017 ms / 455.06 s \\
w/o model reuse & 0.170 / \(10.55^\circ\) / 4.36\% & 0.779 / 0.992 & 100.0\% & $0.006\times$ / 0.017 ms / 5026.94 s \\
w/o identity log & 0.170 / \(10.55^\circ\) / 4.36\% & 0.779 / 0.000 & 100.0\% & $1.52\times$ / 0.018 ms / 455.06 s \\
w/o visibility & 0.336 / \(49.70^\circ\) / 22.06\% & 0.600 / 0.769 & 77.0\% & $1.52\times$ / 0.015 ms / 455.06 s \\
w/o heading stab. & 0.170 / \(66.52^\circ\) / 4.36\% & 0.778 / 0.992 & 100.0\% & $1.52\times$ / 0.017 ms / 455.06 s \\
\bottomrule
\end{tabular}
\end{adjustbox}
\vspace{0.3em}
\begin{minipage}{0.98\textwidth}
\footnotesize Note: State err. T/Y/S denotes Trans./BBox, Yaw MAE, and Scale error. Replay Sp./ID denotes Dynamic Spatial and Dynamic-only ID w/ coverage. Eff. Comp./Upd./Cost denotes compression ratio, state-log update time, and initial construction or replay-materialization cost.
\end{minipage}
\end{table*}

Table~\ref{tab:reuse-ablation} separates accuracy, identity, and efficiency effects. Object-model reuse mainly affects representation efficiency rather than single-frame pose accuracy. Removing it leaves pose and replay scores nearly unchanged, but reduces the compression ratio from $1.52\times$ to $0.006\times$ and increases replay-materialization cost to 5026.94 s. Removing the identity log keeps Dynamic Spatial at 0.779 but reduces Dynamic-only ID w/ coverage from 0.992 to 0.000, showing that frame-wise detections cannot preserve instance identity without an explicit identity log.

Visibility-state maintenance is the most important robustness component under occlusion. Removing it reduces coverage from 100.0\% to 77.0\%, increases normalized translation error from 0.170 to 0.336, increases Yaw MAE from \(10.55^\circ\) to \(49.70^\circ\), and decreases Dynamic Spatial from 0.779 to 0.600. By contrast, removing heading stabilization mainly affects yaw, increasing Yaw MAE to \(66.52^\circ\), while leaving Dynamic Spatial and identity scores nearly unchanged. This indicates that heading stabilization is an orientation-auditing safeguard for cart-like objects rather than the main driver of replay quality. A targeted visualization is provided in Appendix~\ref{app:state-maintenance-visualizations}.

\section{Conclusion}
This paper presented OV-PEAR, an observation-verified framework for pose-scale estimation and 3D reconstruction of unseen industrial manipulation objects. Instead of assuming target CAD models or object-specific training data, OV-PEAR uses generative 3D priors as initial hypotheses and anchors them in the robot frame through appearance-aware view retrieval, explicit scale search, visibility-sensitive registration, and multi-view RGB-D voxel verification. The resulting representation provides an operation-oriented object interface that combines pose, physical scale, contact-relevant geometry, and reusable object identity.

Experiments on public benchmarks, real workbench scenes, connector assembly, and short-term state maintenance validate this design. OV-PEAR achieved an average AR of 0.657 on BOP-Industrial, reached 1.28 cm translation error, 13.17 deg rotation error, and 0.689 voxel IoU in cluttered real workbench scenes, and achieved 99.67\% grasp success and 84.00\% assembly success across USB, HAN10, and RJ45 tasks. The model-library extension further reduced frame-wise state maintenance to a lightweight update after initial object verification. These results show that OV-PEAR is most suitable for industrial workcells where new objects appear but can be repeatedly manipulated or observed after verification. The method is not intended to replace general SLAM, visible-surface mapping, or compliant control, and its current performance still depends on instance segmentation, RGB-D quality, and camera calibration. Future work will focus on active view planning, online model updating, and feedback-guided assembly for more complex dynamic scenes.

\appendices
\section{Fixed Parameters and Runtime Accounting}
\label{app:parameters-runtime}

This appendix reports fixed experimental parameters and deployment-stage runtime accounting. These details support the main evaluation but are separated from the main text to keep the experimental narrative focused on method-level comparisons and task outcomes.

\subsection{Fixed Experimental Parameters}

Table~\ref{tab:app-parameters} lists the fixed experimental parameters used by OV-PEAR. They were set once on the development data according to sensor noise, voxel resolution, number of observation views, and manipulation-tolerance requirements, and were kept unchanged across object categories and test scenes. No test-object-specific or table-result-specific retuning is performed. Variables such as \(N\), \(M\), \(N_i\), and \(N_g\) denote input sizes or data-dependent counts rather than cross-experiment tuning parameters.

\begin{table*}[!t]
	\centering
	\scriptsize
	\caption{Key experimental parameters and fixed rules.}
	\label{tab:app-parameters}
	\begin{adjustbox}{max width=\textwidth}
		\begin{tabularx}{\textwidth}{llXX}
			\toprule
			Symbol & Parameter & Fixed value/rule & Role \\ 
			\midrule
			\([s_{\min},s_{\max}]\) & Scale-search interval & \([0.05,2.5]\) & Covers scale bias from single-view generation. \\
			\(\Delta s\) & Scale step & \(0.025\) & Discretizes scale search under the manipulation-tolerance constraint. \\
			\(\mathcal{M}_v\) & Visible-point selection & \(w_m>0\) & Keeps model points facing the camera. \\
			\(\tau_{\mathrm{corr}}\) & ICP correspondence gate & \(0.015\,\mathrm{m}\) & Removes distant nearest-neighbor matches. \\
			\(\tau_{\mathrm{fit}}\) & Scale-fitness threshold & \(0.015\,\mathrm{m}\) & Scores refined scale candidates. \\
			\(\epsilon_h\) & Depth-hit threshold & \(1.5\) times the voxel side length & Separates depth hits, free-space conflicts, and unknown states. \\
			\(\tau_c\) & Global confidence threshold & \(0.55\) & Selects trusted occupied voxels. \\
			\(\tau_h\) & Minimum depth support & \(1\) & Requires real depth evidence for trusted voxels. \\
			\(\tau_s\) & Predicted support without depth hit & \(2\) & Requires cross-view generative support without a depth hit. \\
			\(N_{\max}\) & Maximum ICP iterations & \(50\) & Bounds each visibility-weighted ICP refinement. \\
			\bottomrule
		\end{tabularx}
	\end{adjustbox}
\end{table*}
\subsection{Deployment Runtime Accounting}

Table~\ref{tab:app-runtime} reports runtime according to deployment stage. The initial construction stage builds an observation-verified object model and writes it into the model library, excluding robot motion, manual annotation, and downstream manipulation execution. The pose-scale update reuses this model and mainly includes view retrieval, scale search, and visibility-sensitive local registration. The continuous-frame state-log update only accounts for object-level bookkeeping after the library has been built, and excludes RGB-D instance segmentation, object modeling, and scene-level SLAM or Gaussian map updating.

\begin{table*}[!t]
	\centering
	\scriptsize
	\caption{Deployment stages and runtime accounting of OV-PEAR.}
	\label{tab:app-runtime}
	\setlength{\tabcolsep}{4pt}
	\renewcommand{\arraystretch}{1.15}
	\begin{adjustbox}{max width=\textwidth}
		\begin{tabularx}{\textwidth}{p{0.20\textwidth}p{0.23\textwidth}p{0.34\textwidth}p{0.17\textwidth}}
			\toprule
			Stage & Trigger & Output or maintained state & Runtime \\
			\midrule
			Object-model construction
			& New object or new library entry
			& Observation-verified object mesh and model-library entry
			& \begin{tabular}[t]{@{}l@{}}
				33.45 s (real workbench)\\
				455.06 s (continuous-frame library)
			\end{tabular} \\
			
			Pose-scale update
			& Before manipulation or when a new instance appears
			& Object pose, scale, and confidence in the robot frame
			& 0.53 s/instance \\
			
			State-log update
			& Each frame after library construction
			& Object ID, pose-scale state, confidence, and visibility log
			& 0.02 ms/frame \\
			\bottomrule
		\end{tabularx}
	\end{adjustbox}
\end{table*}

\section{USB Tolerance and Failure Analysis}
\label{app:usb-failure}

\begin{figure}[!t]
	\centering
	\includegraphics[width=0.9\columnwidth]{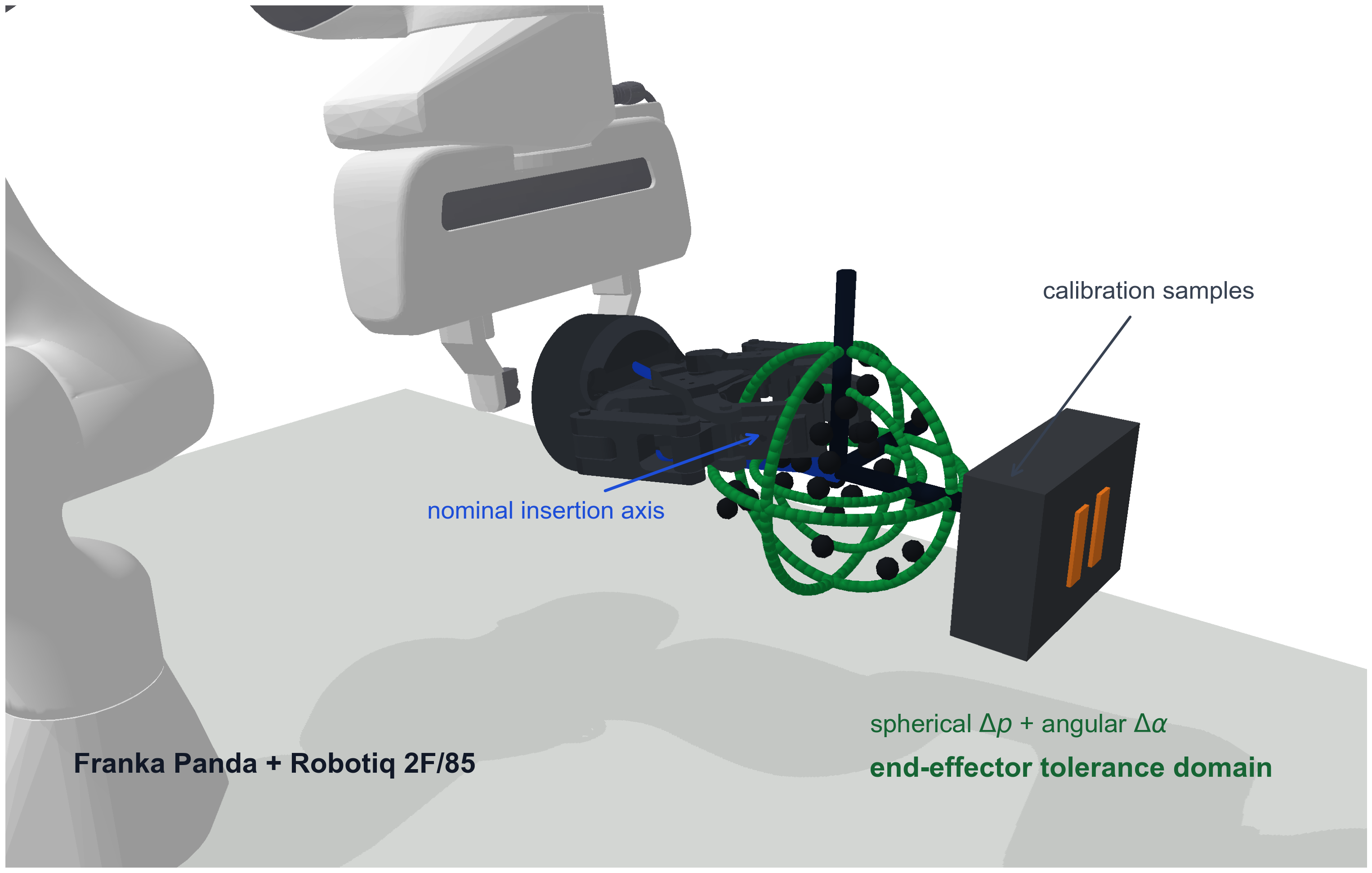}
	\caption{End-effector tolerance calibration for USB compliant insertion. The green spherical region indicates the capture-tolerance region jointly induced by end-effector position and angular perturbations. The blue arrow indicates the nominal insertion direction, and the black samples indicate initial-error samples used for calibration.}
	\label{fig:app-assembly-tolerance}
\end{figure}

To determine the accuracy required for USB insertion, we manually introduce translational and angular offsets and measure the success rate of the compliant insertion strategy. Fig.~\ref{fig:app-assembly-tolerance} illustrates the calibrated capture region, and Table~\ref{tab:app-usb-tolerance} reports the quantitative tolerance range.

\begin{table}[!t]
	\centering
	\scriptsize
	\caption{Tolerance range of USB compliant insertion control with respect to initial pose error.}
	\label{tab:app-usb-tolerance}
	\begin{adjustbox}{max width=\columnwidth}
		\begin{tabularx}{\columnwidth}{p{0.38\columnwidth}cX}
			\toprule
			Initial pose-error range & Success rate & Interpretation \\
			\midrule
			\begin{tabular}[t]{@{}l@{}}
				Translation error \(\leq 3\,\mathrm{mm}\) \\
				Angular error \(\leq 1^\circ\)
			\end{tabular}
			& 100\%
			& Fully inside the stable capture region. \\
			
			\begin{tabular}[t]{@{}l@{}}
				Translation error \(3\)--\(5\,\mathrm{mm}\) \\
				Angular error \(\leq 1^\circ\)
			\end{tabular}
			& 98\%
			& Mostly recoverable with small directional error. \\
			
			\begin{tabular}[t]{@{}l@{}}
				Translation error \(3\)--\(5\,\mathrm{mm}\) \\
				Angular error \(1^\circ\)--\(3^\circ\)
			\end{tabular}
			& 73\%
			& Close to the tolerance boundary and sensitive to contact conditions. \\
			
			\begin{tabular}[t]{@{}l@{}}
				Translation error \(>5\,\mathrm{mm}\) \\
				or angular error \(>3^\circ\)
			\end{tabular}
			& 0\%
			& Outside the capture range of the compliant controller. \\
			\bottomrule
		\end{tabularx}
	\end{adjustbox}
\end{table}

The robot experiments contain two different failure sources that should not be conflated. The first is perception-induced initialization error, which determines whether the target pose enters the capture range of the compliant controller. The second is execution- and contact-induced error, which can occur after a usable target pose has already been provided. These include post-grasp connector shift, hand-eye calibration residuals, robot repeatability, socket-side angular error, and contact-stage jamming.

In the USB initialization comparison, SAM-6D failures are dominated by gross pose outliers, orientation ambiguity, and center bias. Such errors frequently place the connector outside the \(5\,\mathrm{mm}\) and \(3^\circ\) capture range and cannot be recovered by the compliant controller. The ICP-based baseline has fewer outliers when an accurate object model is manually supplied, but it remains vulnerable to degenerate viewpoints and weak local geometric constraints. These cases show that a correct object model alone is insufficient when local registration has a poor convergence basin.

OV-PEAR failures have a different profile. In the single-construction setting, three trials fail because the generated and verified USB mesh lacks sufficient side-view geometry under near top-down observation. Among the remaining usable models, five trials still exceed the calibrated tolerance because small residual translation or angular errors remain outside the compliant-control capture basin. No catastrophic pose outlier is observed. When the verified USB model is reused, all 20 random-initial-pose trials enter the tolerance range, indicating that model verification and reuse remove most perception-side instability.

The end-to-end connector results further show that final assembly failure is not equivalent to front-end perception failure. For USB, HAN10, and RJ45, the object is usually localized and grasped successfully, but insertion can still fail when the connector shifts after grasping or when contact dynamics amplify small pose errors. HAN10 is the most sensitive case because its multi-pin structure strongly penalizes angular deviation. These observations indicate that OV-PEAR provides stable pose-scale initialization, while higher success rates for low-tolerance connectors will require socket-side perception, in-gripper pose estimation, force feedback, or closed-loop visual servoing.

\section{State-Maintenance Visual Details}
\label{app:state-maintenance-visualizations}

\begin{figure}[htbp]
	\centering
	\includegraphics[width=0.9\columnwidth]{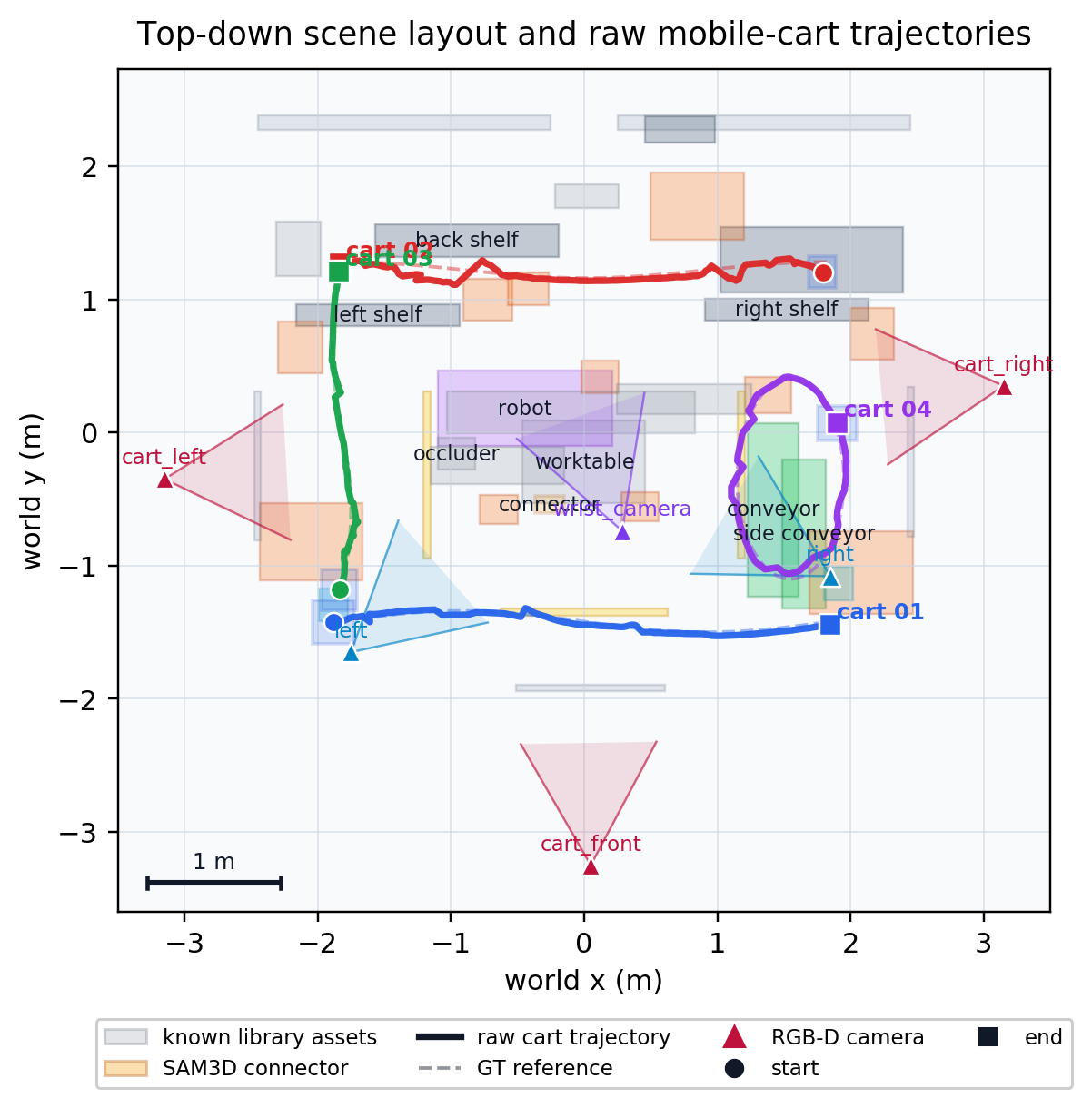}
	\caption{Top-down layout of the short-term state-maintenance scene.}
	\label{fig:app-state-layout}
\end{figure}

\begin{figure}[htbp]
	\centering
	\includegraphics[width=\columnwidth]{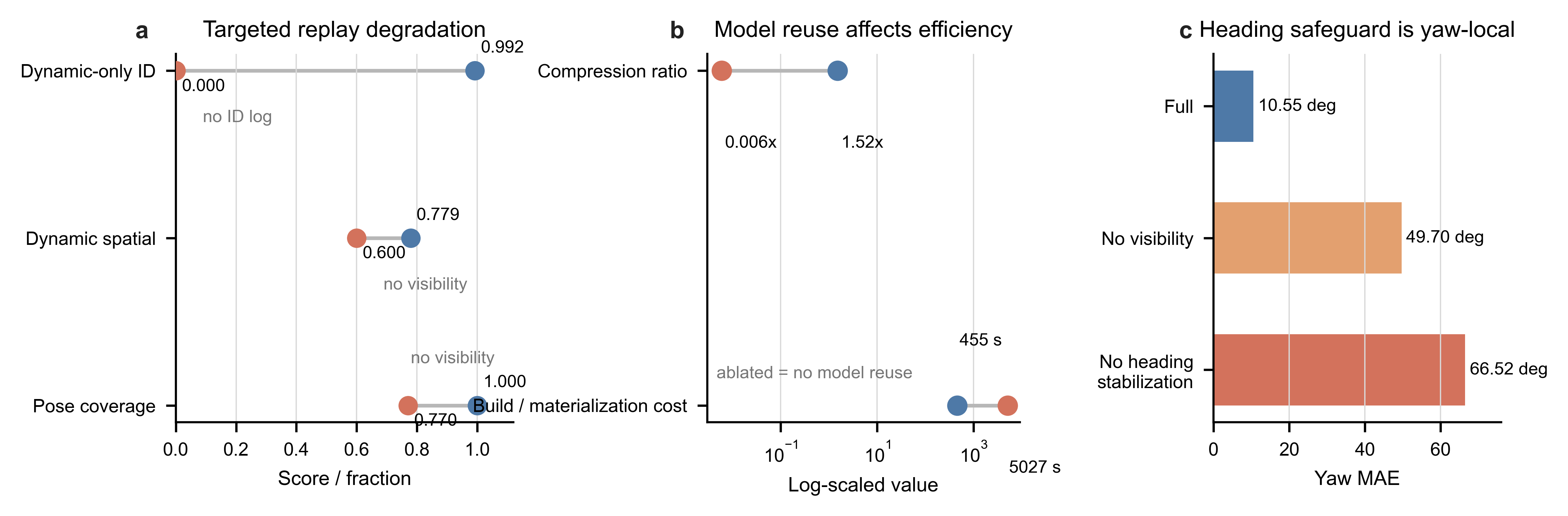}
	\caption{Diagnostic visualization of state-maintenance component ablations.}
	\label{fig:app-reuse-ablation}
\end{figure}

This appendix provides supplementary visual context for the short-term state-maintenance experiment and the state-maintenance component ablation. Fig.~\ref{fig:app-state-layout} shows the registered static assets, RGB-D camera locations, and trajectories of the four moving carts. Fig.~\ref{fig:app-reuse-ablation} visualizes the component ablation summarized in Table~\ref{tab:reuse-ablation}; blue indicates the full OV-PEAR output, whereas orange indicates the corresponding ablation variant.

\clearpage
\bibliographystyle{IEEEtran}
\bibliography{references}

\end{document}